\pdfoutput=1
\documentclass{article}

\usepackage[preprint]{jmlr}
    
\usepackage[utf8]{inputenc} 
\usepackage[T1]{fontenc}    
\usepackage{hyperref}       
\usepackage{url}            
\usepackage{booktabs}       
\usepackage{amsfonts}       
\usepackage{nicefrac}       
\usepackage{microtype}      
\usepackage{amsmath}
\usepackage[export]{adjustbox}
\usepackage{wrapfig}

\usepackage[ruled,vlined]{algorithm2e} 
\usepackage{graphicx}
\usepackage{adjustbox}
\usepackage{makecell}
\usepackage[font=small,labelfont=bf]{caption}
\usepackage{pgfplots}
\pgfplotsset{width=5cm,compat=1.9}
\usepackage{pgfkeys}
\usepackage{siunitx}
\usepackage{enumitem}
\usepackage{bbm}
\usepackage{stmaryrd}

\title{Improving compute efficacy frontiers with SliceOut}

\jmlrheading{1}{2000}{1-48}{4/00}{10/00}{meila00a}{Notin and Gomez, et al.}

\ShortHeadings{SliceOut}{Notin and Gomez, et al.}

\firstpageno{1}

\author{
    \name Pascal Notin \email pascal.notin@cs.ox.ac.uk \\
    \addr University of Oxford\\
    \AND
    \name Aidan N. Gomez \email aidan.gomez@cs.ox.ac.uk \\
    \addr University of Oxford \& Cohere\\
    \AND
    \name Joanna Yoo \email joanna@cohere.ai \\
    \addr Cohere\\
    \AND
    \name Yarin Gal \email yarin@cs.ox.ac.uk \\
    \addr University of Oxford\\
}

\begin{document}

\maketitle

\begin{abstract}
Pushing forward the compute efficacy frontier in deep learning is critical for tasks that require frequent model re-training or workloads that entail training a large number of models. We introduce SliceOut---a dropout-inspired scheme designed to take advantage of GPU memory layout to train deep learning models faster without impacting final test accuracy.
By dropping contiguous sets of units at random, our method realises training speedups through (1) fast memory access and matrix multiplication of smaller tensors, and (2) memory savings by avoiding allocating memory to zero units in weight gradients and activations. At test time, turning off SliceOut performs an implicit ensembling across a linear number of architectures that preserves test accuracy.
We demonstrate 10-40\% speedups and memory reduction with Wide ResNets, EfficientNets, and Transformer models, with minimal to no loss in accuracy. This leads to faster processing of large computational workloads overall, and significantly reduce the resulting energy consumption and CO$_2$ emissions.

\end{abstract}

\section{Introduction}

The success of deep learning over the past two decades has relied heavily on algorithmic and hardware innovations to support ever increasing computational workloads. 
While several methods have been recently introduced to achieve step-improvements in efficacy at inference time (e.g., quantisation, pruning), translating these benefits to training as been a more challenging endeavour given the impact they may have on the training dynamics.
When dealing with a fixed compute budget, the ability to train the same models more rapidly supports shorter research iteration cycles, more extensive hyperparameter or architecture searches, or a reduction in the required energy consumption and the corresponding carbon footprint. In applications that require regular model re-training (e.g., active learning, continual learning), faster training translates into more regular updates and subsequently stronger task performance with the same resources.

\begin{figure*}
    \centering
    \includegraphics[width=0.49\textwidth,valign=t]{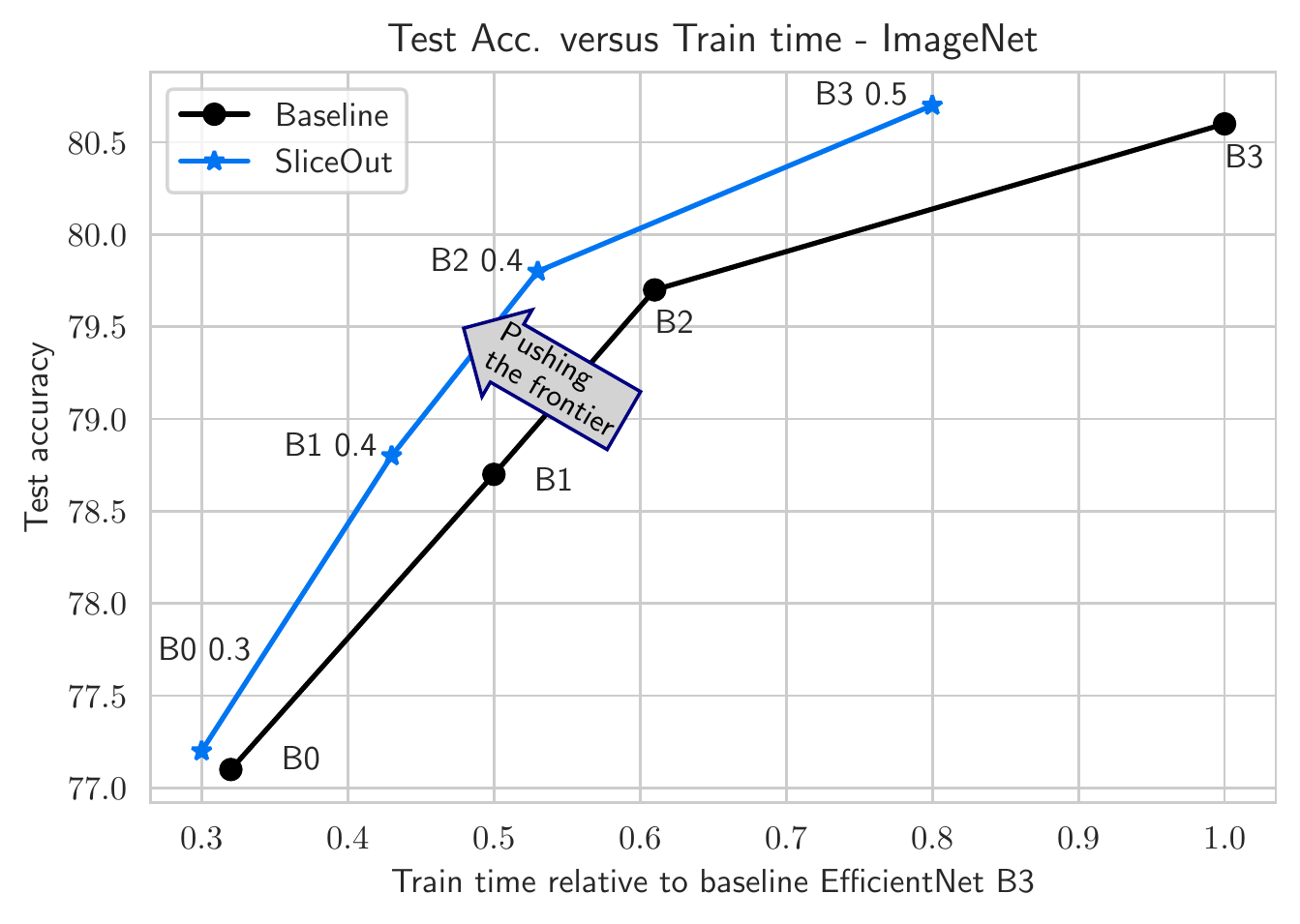}
    \includegraphics[width=0.49\textwidth,valign=t]{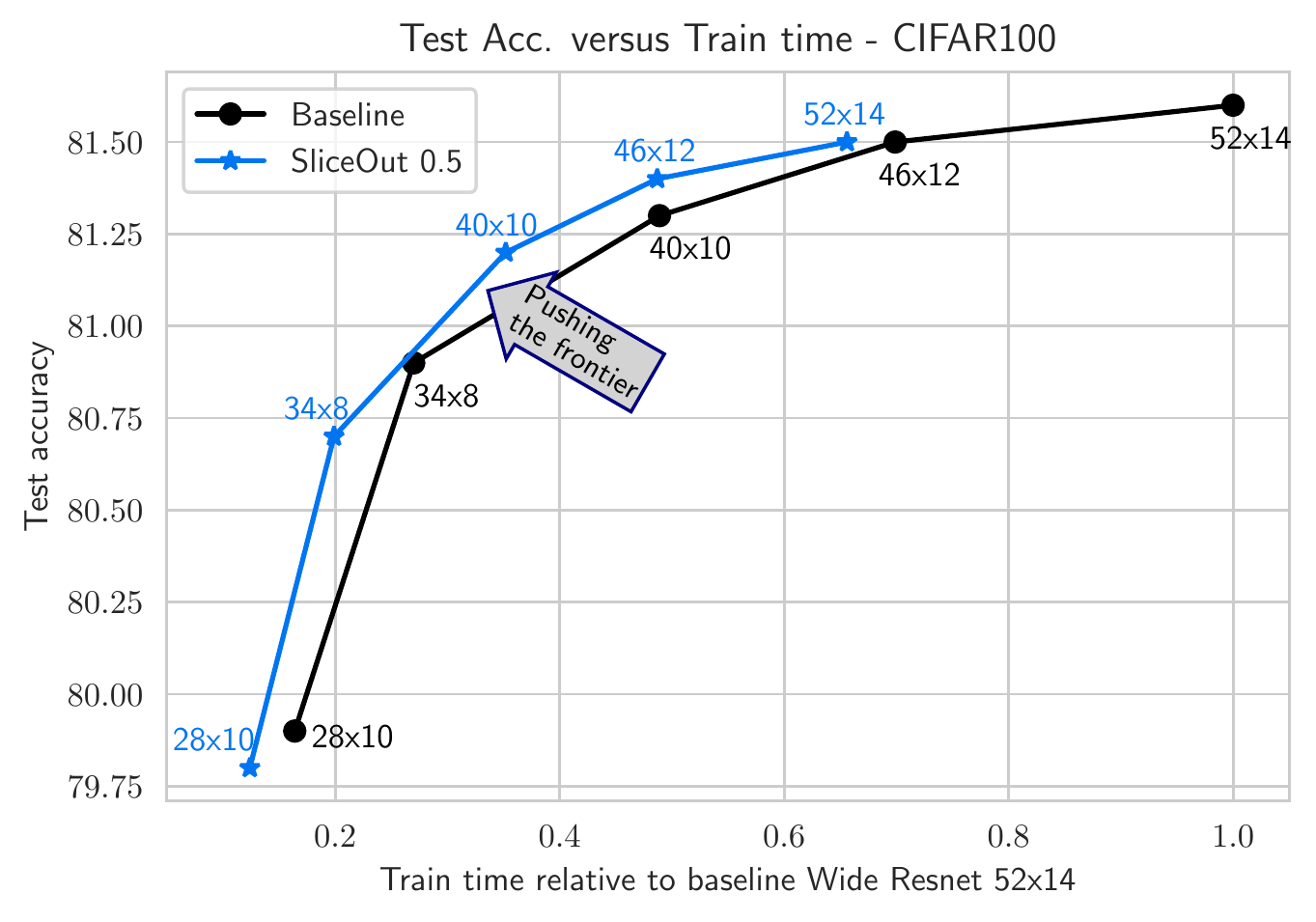}
    \caption{\textbf{Compute efficacy frontiers:} Leveraging SliceOut allows to achieve a more desirable compute efficacy frontier for EfficientNets on Imagenet (left) and Wide ResNets on CIFAR100 (right). Each curve represents the test accuracy Vs required training time (relative to train time of the larger net) for networks trained with SliceOut (blue curves) and without (black curves). See detailed results in tables \ref{Table: WRN - Channel SliceOut & Probabilistic normalisation results} and \ref{Table: EN results - ImageNet}.}
    \label{fig: Compute_efficacy frontiers}
\end{figure*}

In this work, we introduce an architecture-agnostic method to train neural networks faster without compromising on final test accuracy, thereby achieving a more desirable compute efficacy frontier (see Fig.~\ref{fig: Compute_efficacy frontiers}). 

Our proposed method, SliceOut (\S\ref{Section: SliceOut}), draws inspiration from dropout \citep{hinton2012improving,JMLR:v15:srivastava14a}, a regularisation technique widely used in large neural networks. We show that the scheme can be used as an alternative to standard dropout that simultaneously preserves its regularisation benefits while achieving speedups and memory gains at train time.
More generally, we demonstrate it can be also leveraged to achieve training speedups in architectures where no dropout was used in the first place.

SliceOut introduces structure to dropout by slicing contiguous memory segments, i.e., selecting a contiguous range of neighboring neurons and slicing feature tensors or weight matrices row/column-wise (Fig. \ref{fig: SliceOut_scheme}c), as opposed to selecting neurons uniformly at random. From the computational perspective, this strategy takes advantage of GPU memory layout as the operation requires a single access to contiguous memory. From the memory perspective, the zero units, that would physically remain in memory with standard dropout, are removed from memory overhead by the slicing operation. This implies a smaller memory footprint for weight gradients and activations throughout the network, and also results in matrix multiplications with smaller tensors compared to standard dropout. This in turn allows us to fit larger models in memory than would otherwise be possible, or conversely, to train a model of similar size with fewer computing resources. The relative simplicity of the approach as a constrained-form of dropout facilitates its implementation across architectures and deep-learning frameworks. Lastly, SliceOut helps prevent some of the issues that standard dropout has when applied to CNNs (\S\ref{SliceOut and CNNs} and Fig. \ref{fig: SliceOut CNN schemes}).

Our experiments are carried in three settings (\S\ref{Section: Experimental results}): the first consists of relatively small neural networks applied to MNIST and FashionMNIST, to illustrate the benefits of the approach in a simple setting; the second is Wide ResNets applied to CIFAR-10/100 and EfficientNets applied to CIFAR-10/100 and ImageNet, demonstrating significant memory and speedup gains due to the large reduction in ops on the high dimensional feature vectors of CNNs; and the final setting is language modelling with Transformers applied to LM1B, demonstrating the applicability of our method beyond vision tasks.
In all our settings we find that SliceOut performs comparatively (or out-performs) standard dropout in terms of test accuracy, while achieving memory and compute savings of 10-40\%, depending on the model architecture and dropout rate considered. 

Our contributions are as follows:
\begin{itemize}
\item We introduce SliceOut, a general-purpose scheme to train neural networks faster without impacting final test accuracy
\item We derive various sampling and normalisation schemes for the method which preserve (exactly or approximately) the first and second moments of the layers’ output, allowing for efficient deterministic approximations at inference time
\item We implement this new scheme across a diverse set of network architectures - from regular MLPs, to Wide ResNets, EfficientNets and Transformers
\item We quantify the relative speedups and memory gains between the different dropout schemes across experimental setups, demonstrating practical gains with SOTA models with minimal to no impact on test accuracy 
\end{itemize}
\section{Background}
\label{Section: Background}

\subsection{Compute efficacy frontiers}

\begin{wrapfigure}{r}{5.5cm}
    \vspace{-0.5cm}
    \includegraphics[width=2in,page=3]{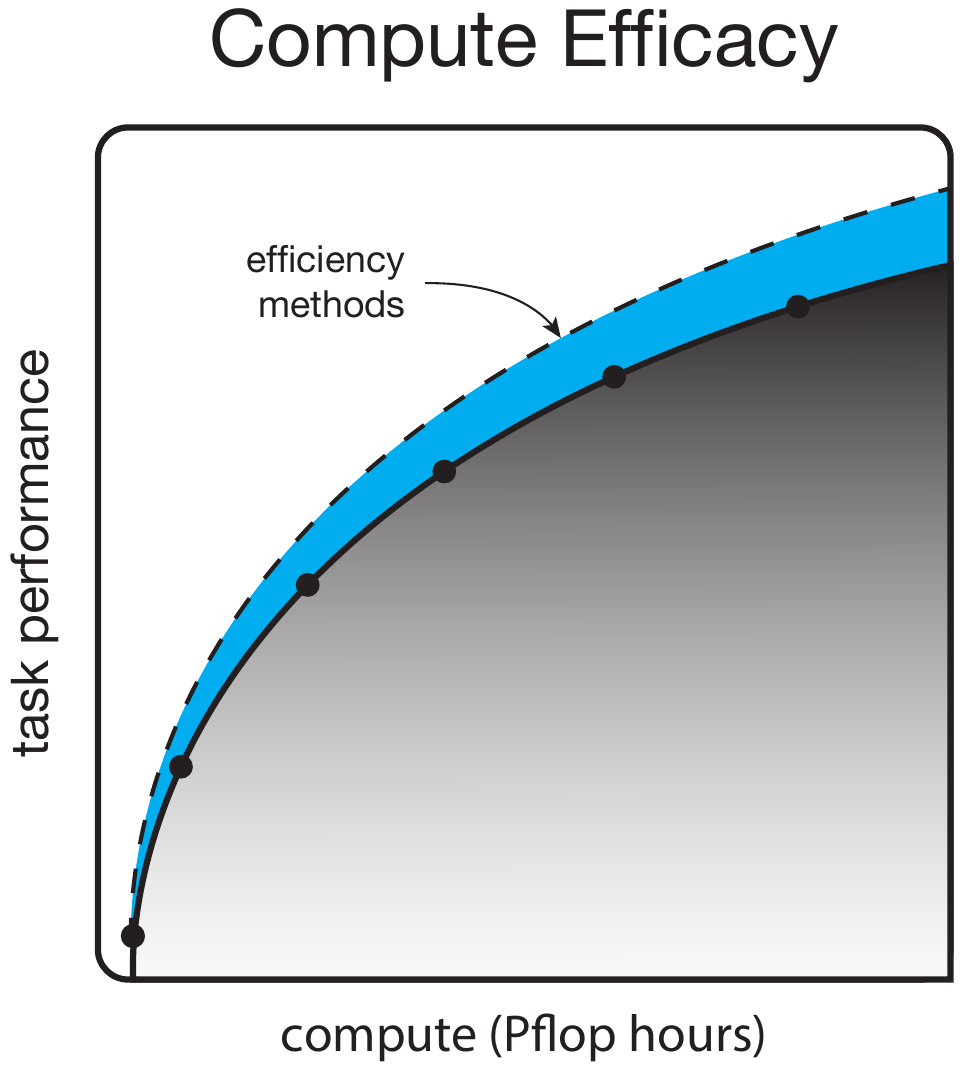}
    \caption{The purpose of developing efficiency in ML is pushing forward the \emph{compute efficacy frontier}: the frontier describing the best model one can obtain using a given amount of compute.}
    \label{fig:eff_frontier}
\end{wrapfigure} 

In the past few years we have observed an unprecedented race to training ever larger neural networks via massive compute resources with the ultimate objective to squeeze in the most parameters possible for a fixed amount of compute -- the latest example being the GPT-3 model with a total of 175 billion parameters \citep{brown2020language}.
Significant progress has also been made towards the ability to train large deep networks very rapidly -- with several teams competing to train high accuracy models on ImageNet in a few minutes \citep{jia2018imagenet4mins, goyal2017largeminibatchImagenet}. \citet{mccandlish2018empirical} investigate the relationship between compute resources and total training time to achieve a fixed test accuracy, and observe Pareto frontiers connecting the two, for example by training a model to solve the Atari Breakout game. 

The aforementioned examples demonstrate the intricate relationships between amount of compute available, overall training duration, and final test accuracy. A convenient way to conceptualise these relationships and describe the objective of progress in ML efficiency is to consider the \emph{compute efficacy frontier}. The compute efficacy frontier  defines the utility of a given amount of compute; that is, given X accelerators operating for Y hours, the frontier describes the best performing model one can obtain (See Figure~\ref{fig:eff_frontier}).
In our work, we introduce a method that pushes the frontier forward by reducing the compute while preserving the task performance. We show that this method is effective across model architectures and across task domains. Importantly, we show that even in highly efficient and optimised settings -- like EfficientNet models for ImageNet (Figure \ref{fig: Compute_efficacy frontiers}) -- our method has a dramatic impact on compute efficacy.

\subsection{Related dropout variants}
Standard dropout randomly ``turns off'' at train time the neurons of a given layer and, implicitly, the weights connected to them. This prevents co-adaptation between neurons \citep{JMLR:v15:srivastava14a}, and empirically results in improved generalisation across a wide range of architectures and tasks \citep{labach2019survey}. 
Standard dropout may also be interpreted as sampling a “thinned” architecture from an exponential number of related networks (\(2^d\) if the layer width is \(d\)) during training, and approximately ensembling these architectures at test time through first-moment propagation \citep{gal2015dropout}.

Since the seminal dropout paper \citep{hinton2012improving}, many alternative dropout schemes have been proposed to improve the efficiency of the technique across a wide range of different neural network architectures. We review the most relevant approaches related to our work.

\begin{figure*}
    \centering
    \includegraphics[width=5.5in]{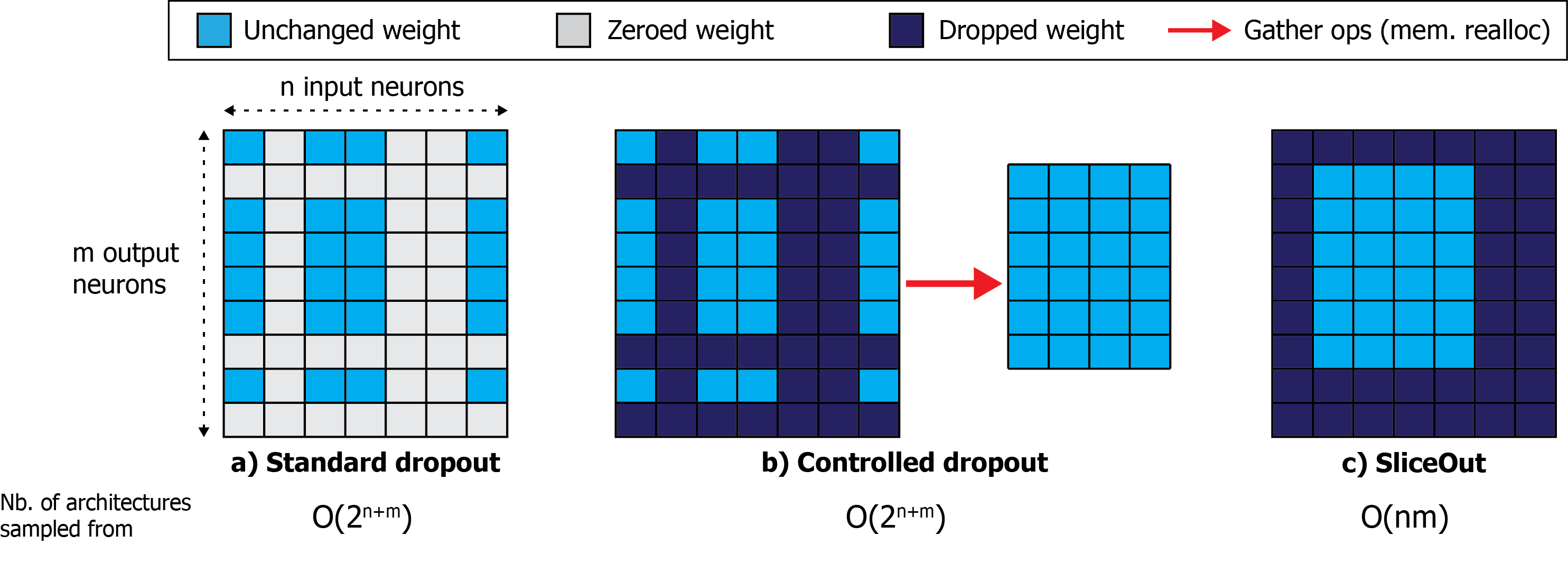}
    \caption{Representation of the effective weight dropout masks for different dropout schemes in a fully connected network. \textbf{a) Standard dropout:} entire rows/columns are set to zero (in practice we typically zero out the output tensor as opposed to the weight) \textbf{b) Controlled dropout:} similar to standard dropout, except non-zero weights are gathered and reallocated in new memory. \textbf{c) SliceOut:} structured weight dropout keeps contiguous set of rows/columns of weight tensors in-place.}
    \label{fig: SliceOut_scheme}
\end{figure*}

\textbf{Standard dropout.} At each training step, the activations from neurons at a layer where dropout is applied are zeroed out with a probability $p$ -- the dropout probability for that layer -- with the forward and backward passes being then performed as usual (Fig. \ref{fig: SliceOut_scheme}a).
During testing, all units of the original architecture are kept to perform the forward pass.
Because a fraction $p$ of units are dropped during training, activations need to be renormalised to preserve the expected value of pre-activations of subsequent layers between train and test, preserving the first and second moments of the layer’s output. 
This normalisation may be performed at test time (``weight scaling inference rule'',  \cite{Goodfellow-et-al-2016}), or during training (``inverted dropout''). The latter is the most popular approach used nowadays and consists of dividing each neuron at a layer where dropout is applied by the probability of it being kept (i.e., divided by $(1-p)$).

\textbf{Controlled dropout.} Controlled dropout \citep{7881693,8122736} was suggested to speed up the training of fully connected networks based on the observation that storing zeroed activations throughout the forward and backward pass leads to computational inefficiencies. The authors propose to keep a random subset of rows or columns of the activation tensors by performing a set of `gather’ operations (\emph{gather ops}) on the corresponding network weights (Fig. \ref{fig: SliceOut_scheme}).  The gather ops select specific weight rows/columns, and allocate new memory into which these rows/columns are copied, so that subsequent multiplications in the forward and backward passes involve smaller tensors.
Although this approach helps avoid unnecessary multiplications, the gather ops’ memory allocations introduce significant overhead. More specifically, the GPU needs to perform a quadratic number of reads and writes in order to create the required reduced tensors. This is not only slow to perform, but also results in duplicating the gathered weight tensors data in memory (Table \ref{Table: Memory requirements dropout schemes}).

\textbf{DropBlock \& SpatialDropout.} Convolutional neural networks require a different scheme than standard dropout to perform effective regularisation \citep{tompson2014efficient,he2015deep}. This is both due to the strong correlations between adjacent pixels present in natural images (and preserved in subsequent feature maps) and the fact convolution kernels operate on nearby pixels. Consequently, when a given pixel is zeroed out, information can still propagate through neighboring pixels as if no dropout had been applied. Several schemes have been proposed to circumvent this limitation, for example by zeroing out contiguous regions of the feature maps \citep{ghiasi2018dropblock} or zeroing out entire convolution filters \citep{tompson2014efficient}.

Further parallels may be drawn between SliceOut and Nested Dropout \citep{rippel2014learning}, in which coherent nested sets of hidden units are dropped in order to learn ordered representations, and with DropEdge \citep{rong2019dropedge}, in which a certain number of edges are removed from the input graph at each training epoch.

\begin{table}[ht]
    \begin{center}
    \caption{\textbf{Comparison of memory usage \& No. of basic operations for different dropout schemes}
    with $b$ the batch size, $n$ \& $m$ the No. of neurons in the input \& output layers resp. \& $p$ the dropout probability applied to both input and output layers (with 0<p<1). \textbf{SliceOut} benefits from the same computation savings as \textbf{Controlled dropout}, without the memory reallocation overhead.}
    \begin{adjustbox}{width=1\textwidth}
    \footnotesize{
    \begin{tabular}{c|ccc}
    \toprule
    Metric & Standard dropout & Controlled dropout & SliceOut \\
    \midrule
    \makecell{No. extra read/writes\\to manipulate weights} & $-$ & $O((1-p)^2*n*m)$ & $O(1)$ \\
    \hline
    \makecell{Extra memory usage\\due to weight copy} & $-$ & $(1-p)^2*n*m$ & $-$ \\
    \hline
    \makecell{No. basic operations\\for weight multiply} & $O(b*n*m)$ & $O((1-p)^2*n*m*b)$ & $O((1-p)^2*n*m*b)$\\
    \hline
    \makecell{Size of output\\activations tensor} & $m*b$ & $(1-p)*m*b$ & $(1-p)*m*b$\\
    \bottomrule
    \end{tabular}
    }
    \end{adjustbox}
    \label{Table: Memory requirements dropout schemes}
    \end{center}
\end{table}

\section{SliceOut}
\label{Section: SliceOut}
SliceOut is a structured weight dropout scheme aimed at speeding up computations and reducing cached memory footprint, while preserving the regularisation benefits of standard dropout.
We first convert the dropout rate into an expected number of nodes that should be kept at a layer where SliceOut is applied, i.e. the “slice width”.
During training, we uniformly sample the starting index of the slice (restricting to a subset of eligible positions, as detailed in Appendix~\ref{Appendix: Constraints on eligible positions for slicing}), then ``slice'' (see next paragraph) the relevant rows and columns of the weights and biases that precede / follow the layer(s) where SliceOut is applied (Fig. \ref{fig: SliceOut_scheme}).
We then perform the forward and backward passes with the sliced weights and biases, updating the corresponding slice(s) of the original weight matrices in-place. We repeat this end-to-end process, sampling different slices at each step, until convergence (Algorithm \ref{Algorithm: SliceOut}).
At test time, we use the full network without dropping any weights or biases, similar to standard dropout.

\subsection{The slice op}
Slicing is a fast and memory efficient operation: it selects the tensor elements of interest with a single memory access, and performs tensor operations with the logical tensors in-place \citep{harris2020array,Pytorch_NEURIPS2019_9015}. The slice operation (\emph{slice op}) only changes the logical view into the memory, but not the physical memory. When a GPU matmul or conv kernel (both GPU functions) is called, it only sees the weights within that view, and does its operation with those weights without having to move anything in memory.
SliceOut enjoys speed-ups from performing forward and backward passes with smaller tensors; Furthermore, as we now need only keep the smaller sliced activation tensors in memory to perform the backward pass at train time, we save on activation storage.

At test time we use the full network, and therefore there is no difference in memory usage to a network trained with standard dropout. However, the memory bottleneck for large networks is typically at train time since we are required to store intermediate activations to compute gradients on the backward pass.

\subsection{Normalisation}\label{normalisation}
After applying dropout, it is necessary to re-normalise activations in order to preserve the moments of their distributions and avoid the network outputs exploding or collapsing to zeros. We experimented with different approaches to normalise activations after dropout, and describe here the two that lead to the best results in experimental settings (Appendix \ref{Appendix: Activation normalization & moment preservation}):
\begin{itemize}[noitemsep,topsep=0pt]
    \item \textbf{Flow normalisation:} We divide activations by the \emph{expected proportion} of nodes kept at that layer during training (i.e., the ratio of the slice width to the full layer width). Intuitively, this helps keep constant the expected values of pre-activations at subsequent layers.
    \item \textbf{Probabilistic normalisation:} We divide each node by the probability of \emph{this specific node} being kept during training. This helps ensure that, on average during training, the activations stemming from this particular node are equal to what they would be at test time.
\end{itemize}

These two normalisations coincide in the standard dropout case, where the expected proportion of nodes kept at a given layer is exactly equal to the probability of each node to be kept during training. This is not the case in SliceOut, as we impose constraints on eligible slices during sampling to avoid memory re-allocations and keep the size of tensors constant throughout training (Appendix~\ref{Appendix: Constraints on eligible positions for slicing}): nodes around the edges are less likely to be selected at a given training step.

\subsection{Regularisation and ensembling}
While standard dropout samples a “thinned” network from an exponential number of possible architectures, SliceOut samples from a linear or quadratic number of architectures.\footnote{If SliceOut is applied at only one layer, we only take slices row-wise of the corresponding weight vector (and column-wise of the subsequent weight vector), thereby sampling from a linear number of architectures. If SliceOut is applied at two consecutive layers, we slice the second weight matrix row and column wise, thereby sampling from a quadratic number of architectures (Appendix~\ref{Appendix: Number of architectures sampled from}).}
As a result SliceOut can be seen as a milder regularisation scheme (for a fixed dropout probability value). We observe in several experimental settings that, beyond a certain dropout probability threshold, the performance drops more sharply in standard dropout than in SliceOut. This increased stability makes SliceOut less sensitive to the chosen dropout probability, enabling higher drop rates.

\begin{algorithm}[H]
\caption{Slice dropout algorithm - Simple FFN with L hidden layers} 
\footnotesize
    Let $W_l$, with $l \in [1-L]$, be the weights tensor of the $l^{th}$ hidden layer\;
    Let $f_l(.)$ be the non-linearity applied at the $l^{th}$ layer\;
    \For {$training\_step \gets 1\ to\ T$}{
        Sample mini-batch $(x,y)$\;
        \For {$layer_l \gets 1\ to\ L$}{Sample slice: $Slice_l=(start_l,end_l)$\;}
	    \For {$layer_l \gets 1\ to\ L$}{Slice weights:
	        $W_l\_slice$ = $W_l[(start_l,end_l),(start_{l-1},end_{l-1})]$ -- where $(start_0,end_0)$\ selects\ the\ full\ input\;
	    }
	    Perform forward pass with sliced weights:
	    \For {$layer_l \gets 1\ to\ L$}{$x \gets f_l(norm(W_l\_slice \cdot x))$ -- where norm(.) is the activation normalisation applied post dropout}
	    Perform backward pass with sliced weights\;
    }
    \label{Algorithm: SliceOut}
\end{algorithm}

\subsection{SliceOut and CNNs}
\label{SliceOut and CNNs}
\begin{figure*}[h]
    \centering
    \includegraphics[width=4.5in]{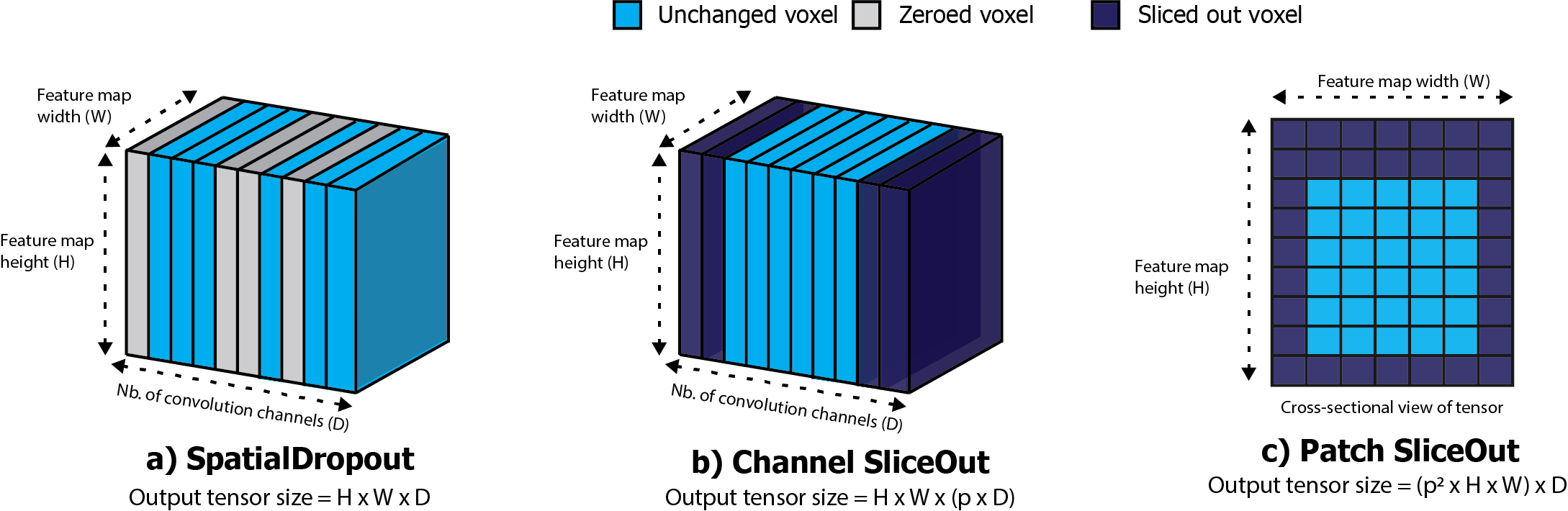}
    \caption{Comparison of the feature tensor of a convolution layer where different dropout schemes are applied \\
    \textbf{a) SpatialDropout:} randomly sets entire convolution channels to zero. \textbf{b) Channel SliceOut} randomly selects a contiguous set of convolution channels, resulting in a more compact feature tensor (other channels are never allocated in memory) \textbf{c) Patch SliceOut:} selects a contiguous block of the input tensor across feature maps, then performs the convolution.}
    \label{fig: SliceOut CNN schemes}
\end{figure*}

Our SliceOut schemes for CNNs (Fig.~\ref{fig: SliceOut CNN schemes}) draw inspiration from the prior dropout schemes tailored to CNNs discussed in \S\ref{Section: Background} \citep{tompson2014efficient,ghiasi2018dropblock}:
\begin{itemize}[noitemsep,topsep=0pt]
    \item \textbf{Channel-SliceOut}: slicing contiguous sets of channels for a given convolution kernel
    \item \textbf{Patch-SliceOut}: slicing contiguous 2D chunks of the input activation tensors, and then performing the convolution
\end{itemize}

Channel-SliceOut builds on the SpatialDropout scheme \citep{tompson2014efficient}, with the critical difference that we directly slice the convolution kernels instead of zeroing out feature maps of the output activation tensor. This results in smaller output tensors and helps avoiding performing tensor operations for which the outcome will be ultimately be set to zero.
Patch-SliceOut can be seen as performing the complement operation to what is done in Cutout \citep{devries2017improved} (on the input image), or more generally in DropBlock \citep{ghiasi2018dropblock}, where units in a contiguous region of a feature map are dropped together, except that we slice out zeros instead of carrying them around.

\subsection{SliceOut and Transformers}

\begin{figure*}[ht]
    \centering
    \vbox{
    \begin{minipage}[t]{0.49\textwidth}
        \centering
        Attention Mechanism \vspace{3mm}\\
        \includegraphics[height=5cm]{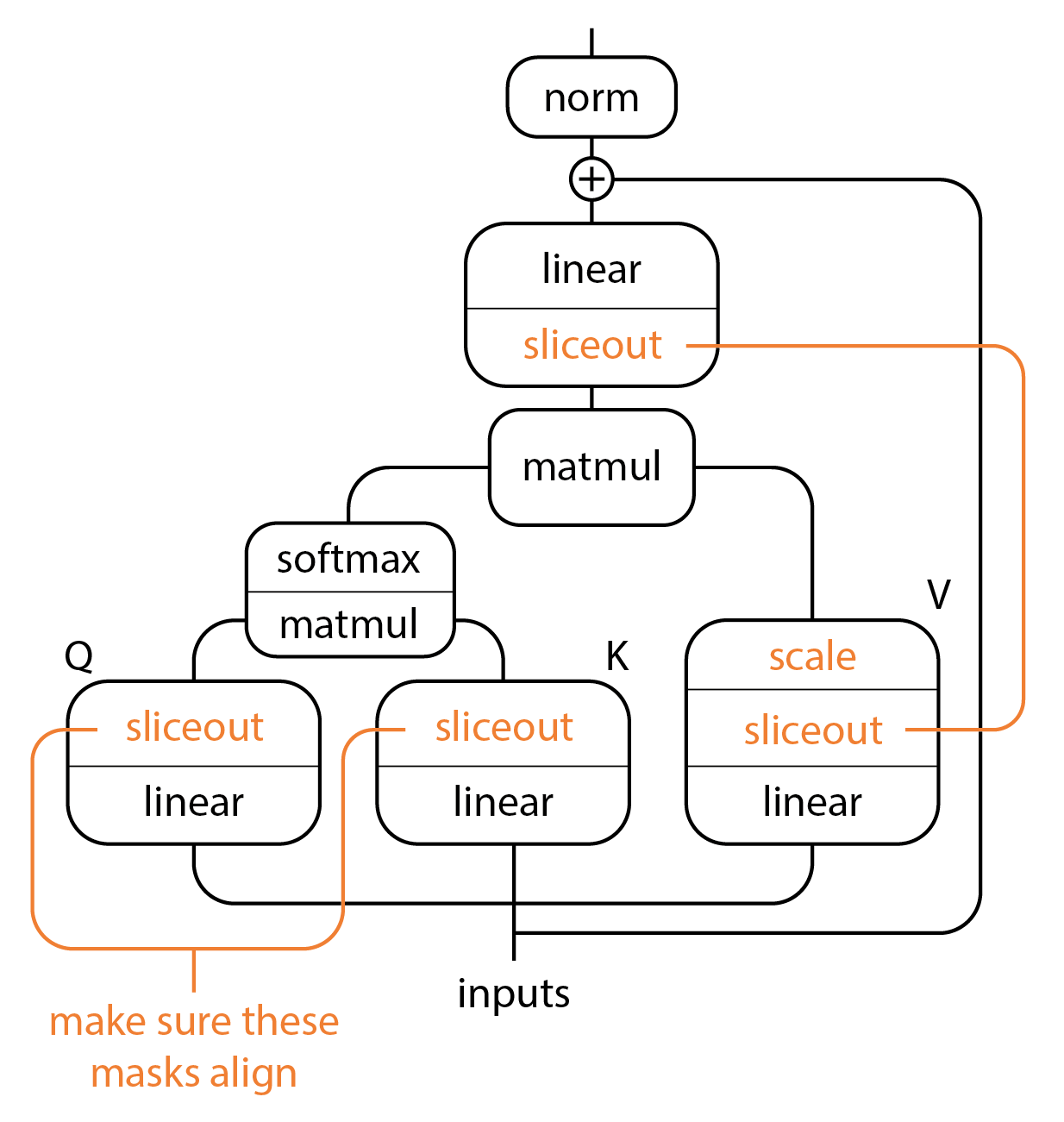}
    \end{minipage}
    \begin{minipage}[t]{0.49\textwidth}
        \centering
        Feed-Forward Network \vspace{3mm}\\
        \includegraphics[height=5cm]{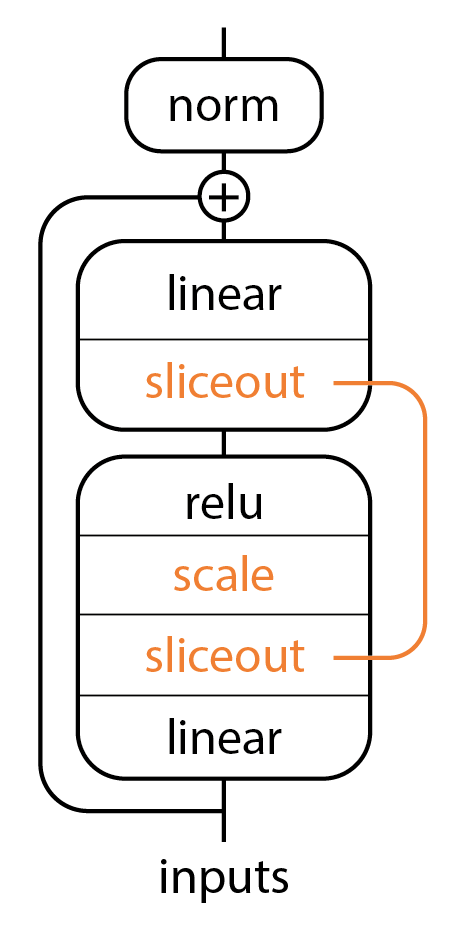}
    \end{minipage}
    }
    \caption{Transformer architecture with SliceOut}
    \label{fig: SliceOut Transformers}
\end{figure*}

Transformers \citep{vaswani2017attention} represent the state of the art across a host of natural language benchmarks and have seen adoption across academia and industry. One short-coming of the architecture is the considerable memory requirements demanded by the model architecture since Transformers tend to improve their performance dramatically with the number of parameters they are given. This observation has lead to several strategies to construct larger and better models (a quick overview of the Transformer architecture is given in Fig.~\ref{fig: SliceOut Transformers}).

SliceOut represents a complementary technique to the standard model-scaling measures taken in the literature (e.g., distributed data and model-parallelism, memory efficiency-focused optimisers \citep{shazeer2018adafactor}) and can be used in conjunction with them.

In our implementation of SliceOut in Transformers we do not normalise the queries and keys as in \S\ref{normalisation}. Instead, we modify the temperature value (\(\alpha\)) used in the attention weights:
\[W_{attn}=\operatorname{softmax}{\left(\frac{QK^\top}{\sqrt{\alpha}}\right)}\]
In a Transformer \(\alpha\) is generally set to the dimensionality of the vectors in the queries and keys, but in our case, SliceOut changes the dimensionality of those vectors during training, and so we adjust \(\alpha\) to be the new dimensionality of these vector after SliceOut. 
We do still perform normalisation (\S\ref{normalisation}) on the values and within the feed-forward networks (Fig.~\ref{fig: SliceOut Transformers}; Note: In the figure, normalisation is denoted ``scale'' while ``norm'' refers to layer normalisation, as in the original Transformer paper).

Since there is a dot product taken between each of the queries and keys, it is necessary that the sliced out indices of those vectors are aligned. That is, SliceOut slices out some contiguous set of elements from a query vector \(Q_{\text{sliced}}=(q_i, \dots, q_{q+d})\); it is, of course, extremely important than these indices are the same for the sliced keys \(K_{\text{sliced}}=(k_i, \dots, k_{q+d})\). Similarly, when slicing weight matrices we must ensure that the slices made along the leading dimension align with the slices applied to the incoming activation vector. See the orange lines in Fig.~\ref{fig: SliceOut Transformers} for a pictorial description of indices that must be aligned.
\section{Experimental results}
\label{Section: Experimental results}

We quantify the benefits of SliceOut across several neural network architectures and application domains:  fully connected networks on MNIST and FashionMNIST datasets (\S\ref{Experiment: FCN}), Wide ResNets on the CIFAR-10 and CIFAR-100 datasets (\S\ref{Experiment: WRN}, EfficientNets on CIFAR-10/100 and ImageNet (\ref{Experiment: EN}), and Transformers on the LM1B dataset (\S\ref{Experiment: Transformers}). For each experiment, we train the different networks until convergence, measure speedups based on the train time per epoch, and memory gains via the maximum GPU memory managed by the caching allocator at each epoch. All details about the training procedure and hyperparameters used are provided in Appendix~\ref{Appendix: Experimental details}, including comparisons between the different sampling and normalisation schemes introduced in \S\ref{Section: SliceOut}.

\subsection{Fully connected networks}
\label{Experiment: FCN}
This first set of experiments is performed in a simpler setting aimed at studying the properties of our method with fully connected networks on the MNIST \citep{Lecun1998MNIST} and FashionMNIST \citep{xiao2017fashionmnist} datasets.

\begin{figure*}

\begin{tikzpicture}
\begin{axis}[
    title={a. Cached GPU memory (\% of standard)},
    xlabel={Dropout rate},
    xmin=0, xmax=0.5,
    ymin=50, ymax=110,
    xtick={0,0.1,0.2,0.3,0.4,0.5},
    ytick={50,60,70,80,90,100,110},
    legend pos=south west,
    ymajorgrids=true,
    grid style=dashed,
    title style={font=\tiny},
    label style={font=\tiny},
    tick label style={font=\tiny},
    legend style={font=\tiny},
]
\addplot[color=red,mark=square,]
    coordinates {
    (0,100)(0.1,100)(0.2,100)(0.3,100)(0.4,100)(0.5,100)
    };
\addplot[color=blue,mark=o,]
    coordinates {
    (0,100)(0.1,93.2)(0.2,87)(0.3,81.5)(0.4,76.6)(0.5,73)
    };
\legend{Standard, SliceOut}
\end{axis}
\end{tikzpicture}
\hskip 5pt
\begin{tikzpicture}
\begin{axis}[
    title={b. Train time per epoch (\% of standard)},
    xlabel={Dropout rate},
    xmin=0, xmax=0.5,
    ymin=50, ymax=110,
    xtick={0,0.1,0.2,0.3,0.4,0.5},
    ytick={50,60,70,80,90,100,110},
    legend pos=south west,
    ymajorgrids=true,
    grid style=dashed,
    title style={font=\tiny},
    label style={font=\tiny},
    tick label style={font=\tiny},
    legend style={font=\tiny},
]
\addplot[color=red,mark=square,]
    coordinates {
    (0,100)(0.1,100)(0.2,100)(0.3,100)(0.4,100)(0.5,100)
    };
\addplot[color=blue,mark=o,]
    coordinates {
    (0,100)(0.1,98.9)(0.2,95.3)(0.3,93.4)(0.4,90.5)(0.5,87.3)
    };
\legend{Standard, SliceOut}
\end{axis}
\end{tikzpicture}
\hskip 5pt
\begin{tikzpicture}
\begin{axis}[
   title={c. Test accuracy Vs No. of epochs},
    xlabel={No. of epochs},
    xmin=0, xmax=100,
    ymin=75, ymax=92,
    xtick={0,20,40,60,80,100},
    ytick={75,80,85,90},
    legend pos=south west,
    ymajorgrids=true,
    grid style=dashed,
    title style={font=\tiny},
    label style={font=\tiny},
    tick label style={font=\tiny},
    legend style={font=\tiny},
]
\addplot[color=red,]
    coordinates {
    (1,78.1)	(2,81.25)	(3,83.27)	(4,83.92)	(5,84.34)	(6,84.67)	(7,84.88)	(8,85.59)	(9,85.76)	(10,85.95)	(11,86.19)	(12,86.4)	(13,86.34)	(14,86.83)	(15,86.92)	(16,87.01)	(17,86.94)	(18,87.15)	(19,87.21)	(20,87.03)	(21,87.32)	(22,87.21)	(23,87.63)	(24,87.49)	(25,87.52)	(26,87.64)	(27,87.55)	(28,87.49)	(29,87.55)	(30,87.7)	(31,87.81)	(32,87.95)	(33,87.84)	(34,87.8)	(35,88.03)	(36,87.86)	(37,87.6)	(38,87.65)	(39,88.01)	(40,87.91)	(41,87.96)	(42,88.08)	(43,87.95)	(44,88.1)	(45,87.97)	(46,88.04)	(47,88.12)	(48,87.93)	(49,88.01)	(50,88.13)	(51,88.07)	(52,87.98)	(53,88.01)	(54,87.83)	(55,87.96)	(56,88.25)	(57,88.19)	(58,87.87)	(59,88.18)	(60,88.22)	(61,88.04)	(62,88.3)	(63,87.92)	(64,88.01)	(65,88.06)	(66,88.33)	(67,88.31)	(68,88.44)	(69,88.14)	(70,88.31)	(71,88.31)	(72,88.2)	(73,88.15)	(74,87.99)	(75,88.42)	(76,88.32)	(77,88.29)	(78,88.25)	(79,88.29)	(80,88.36)	(81,87.85)	(82,88.26)	(83,88.32)	(84,88.43)	(85,88.26)	(86,88.24)	(87,88.16)	(88,88.25)	(89,88.14)	(90,88.48)	(91,88.32)	(92,88.25)	(93,88.58)	(94,87.89)	(95,88.12)	(96,88.36)	(97,88.34)	(98,88.1)	(99,88.13)	(100,88.33)
    };
\addplot[color=blue,]
    coordinates {
    (1,77.69)	(2,81.65)	(3,83.87)	(4,84.75)	(5,85.13)	(6,85.56)	(7,85.78)	(8,86.42)	(9,86.68)	(10,86.52)	(11,87)	(12,87.07)	(13,87.09)	(14,87.48)	(15,87.33)	(16,87.58)	(17,87.5)	(18,87.61)	(19,87.71)	(20,87.74)	(21,88.01)	(22,87.98)	(23,88.02)	(24,88.29)	(25,88.11)	(26,88.2)	(27,88.3)	(28,88.23)	(29,88.19)	(30,88.29)	(31,88.32)	(32,88.5)	(33,88.6)	(34,88.59)	(35,88.5)	(36,88.63)	(37,88.49)	(38,88.45)	(39,88.71)	(40,88.5)	(41,88.52)	(42,88.61)	(43,88.45)	(44,88.9)	(45,88.71)	(46,88.62)	(47,88.73)	(48,88.77)	(49,88.78)	(50,88.93)	(51,88.79)	(52,88.78)	(53,88.72)	(54,88.85)	(55,88.92)	(56,88.88)	(57,88.81)	(58,88.86)	(59,88.93)	(60,88.9)	(61,88.87)	(62,88.95)	(63,88.82)	(64,88.9)	(65,88.99)	(66,89.06)	(67,88.84)	(68,89.09)	(69,89.15)	(70,88.98)	(71,89.07)	(72,88.67)	(73,89.06)	(74,88.9)	(75,88.77)	(76,89.06)	(77,88.99)	(78,88.9)	(79,88.95)	(80,89.16)	(81,88.99)	(82,89.07)	(83,89.26)	(84,89.22)	(85,89.01)	(86,88.9)	(87,89.15)	(88,88.93)	(89,89.21)	(90,89.14)	(91,89.19)	(92,89.11)	(93,89.18)	(94,88.91)	(95,89.35)	(96,89.25)	(97,89.27)	(98,89.24)	(99,89.38)	(100,89.44)
    };
\legend{Standard $p=0.5$, SliceOut $p=0.5$}
\end{axis}
\end{tikzpicture}
\caption{FashionMNIST - We achieve ~$30\%$ memory savings (a.) and ~$15\%$ training speedups (b.) from replacing standard dropout with SliceOut in a simple fully connected network with 3 hidden layers, converging slightly faster and to a higher test accuracy value (c., here for dropout rate = $0.5$, although similar trends were observed for any dropout rate $\leq 0.5$). Results were averaged over 4 independent runs (see Appendix~\ref{Appendix: FashionMNIST})}
\label{Fig: FashionMNIST main results}
\end{figure*}
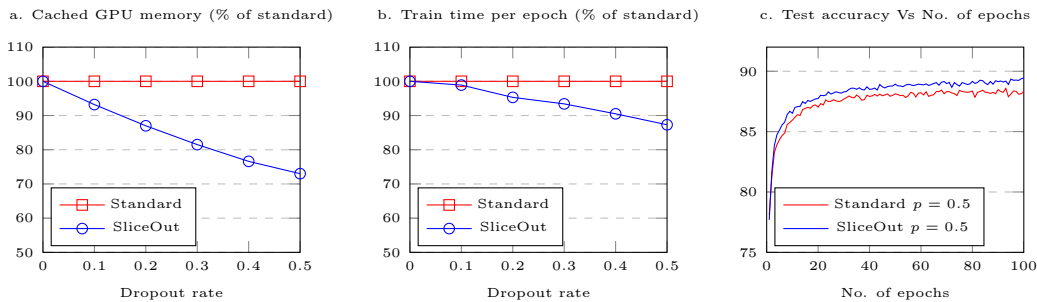

In the FashionMNIST experiments, we observe not only speedups (up to 15\%) and cached GPU memory savings (up to 30\%) with SliceOut, we also converge faster and to a higher test accuracy value (Fig. \ref{Fig: FashionMNIST main results}) when typical dropout rates are applied (i.e., $p\leq 0.5$). The highest test accuracy obtained with SliceOut across all hyperparameter settings tested was $90.0 \pm \SI{0.03}{\percent}$ (obtained with $p=0.1$), while the highest with standard dropout (also for $p=0.1$) was $89.6 \pm \SI{0.08}{\percent}$ (no dropout lead to a test accuracy similar to the latter, see Appendix~\ref{Appendix: FashionMNIST}).

In the MNIST experiments, we observe similar speedups and memory gains from SliceOut, although there was no statistically significant difference in terms of top test accuracy (Appendix~\ref{Appendix: MNIST}).

Across both experiments, controlled dropout was converging to similar test accuracy values, but was systematically slower and less memory efficient than SliceOut (Appendix~\ref{Appendix: Experimental details}).

\subsection{Wide ResNets}
\label{Experiment: WRN}
Wide ResNets \citep{zagoruyko2016wide} are a variant of the original ResNet architecture that achieve higher test accuracy by simultaneously reducing the depth of the network and increasing the number of convolution filters in each residual block. The architecture strings together several ``Wide-dropout'' blocks, progressively increasing the number of channels and reducing the height \& width of the activation tensors. Standard dropout is used critically in each residual block between the two $3\text{x}3$ convolutions, to prevent potential overfitting resulting from the channel widening. 

\begin{figure*}[!ht]
    \centering
    \includegraphics[width=2.5in]{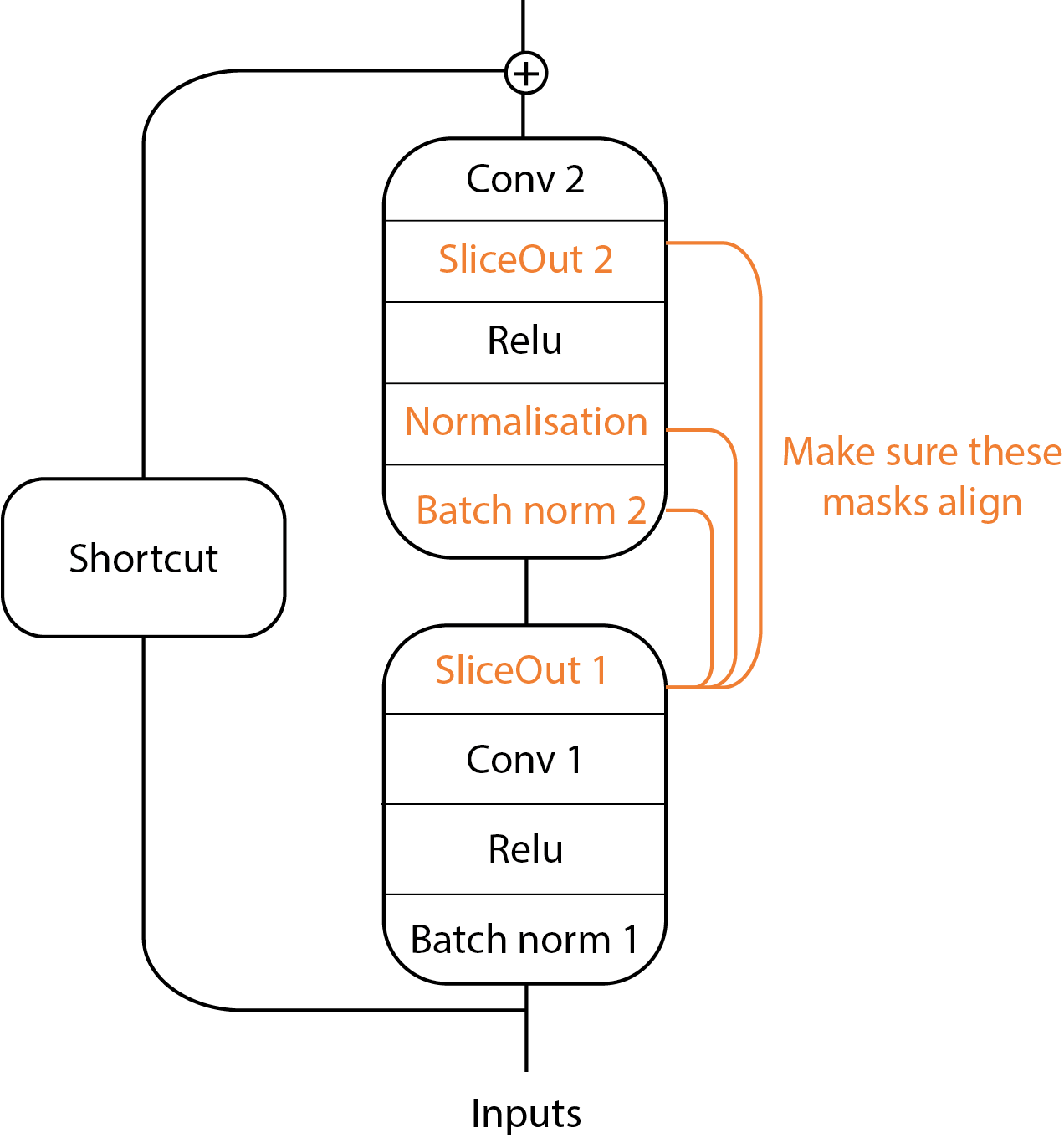}
    \hspace{1 cm}
    \includegraphics[width=2.5in]{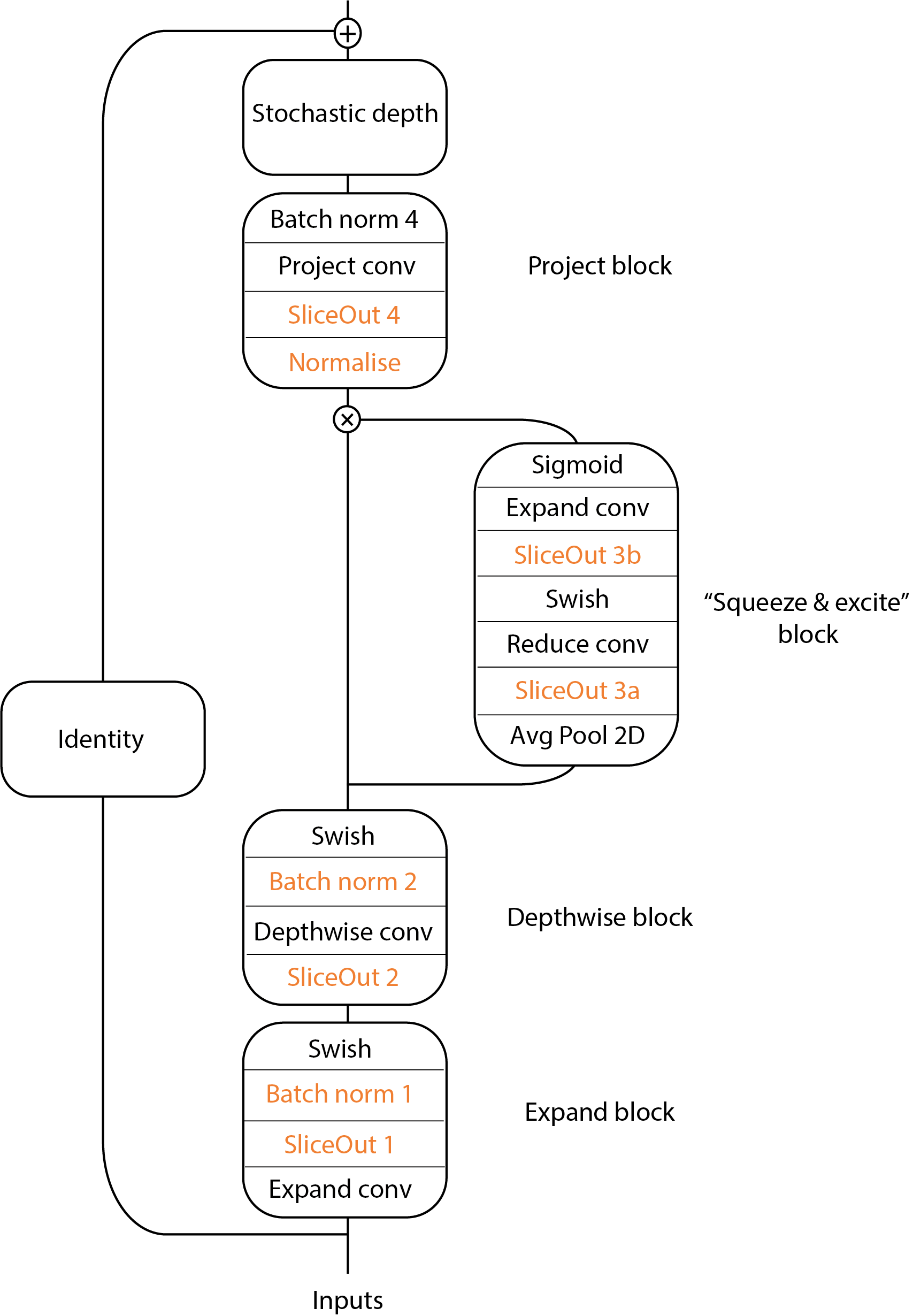}
    \caption{\textbf{Wide ResNet residual block and EfficientNets MBConv block with SliceOut.} The selected slices for the items in orange need to be aligned for a given forward/backward pass.}
    \label{WRN - Architecture diagram}
    \label{EN - Architecture diagram}
\end{figure*}

We remove the standard dropout layer in the original ``Wide-dropout'' block and experiment with our two SliceOut schemes for CNNs (see Fig.~\ref{fig: SliceOut CNN schemes} and architecture diagram on Fig.~\ref{WRN - Architecture diagram}):
\begin{itemize}[noitemsep,topsep=0pt]
    \item \textbf{Channel-SliceOut:} we apply SliceOut on the first $3\text{x}3$ convolution across all residual blocks. It is critical to ensure that we operate on the same slice at the subsequent convolution layer, and the batch norm in-between;
    \item \textbf{Patch-SliceOut:} we apply Patch-SliceOut on the input tensor to the first $3\text{x}3$ convolution, across all blocks.
\end{itemize}
For both schemes, performing the normalisation after the second batch norm and right before the final projection convolution (``delayed normalisation'', Fig.~\ref{WRN - Architecture diagram}) helps further increase test accuracy. We observe higher performance when using the Probabilistic normalisation scheme over the Flow normalisation (\S\ref{normalisation}), and for Channel-SliceOut over Patch-SliceOut (see Appendix~\ref{Appendix: WRN}).

When using SliceOut across a range of Wide Resnet architectures on CIFAR-10/100 (Table.~\ref{Table: WRN - Channel SliceOut & Probabilistic normalisation results}), we obtain training speedups of up to $35\%$ and memory gains of up to $25\%$ with no impact on test accuracy. This translates into a superior compute efficiency frontier (Fig.~\ref{fig: Compute_efficacy frontiers}). For example, we are able to train a 46x12 architecture with SliceOut as fast as a 40x10 architecture without SliceOut, and achieve a higher test accuracy as a result.

\begin{table}[!ht]
\begin{center}
\caption{\textbf{Wide ResNets results.} Training time \& Max cached GPU memory are respectively the relative train time speedups per epoch for a network trained with SliceOut Vs standard dropout, and the maximum cached GPU memory during training. Results are averaged over 5 independent runs. Reported baseline values (standard dropout) are obtained via an hyperparameter search over dropout rates and selecting the value yielding the highest test accuracy. SliceOut results are obtained with a 0.5 rate, Channel-SliceOut and Probabilistic normalisation.}

\begin{adjustbox}{width=0.9\textwidth}
\footnotesize{
\begin{tabular}{cc|c|ccc}
\toprule
Dataset & Architecture & Test accuracy & Test accuracy & Training & Max cached \\
 & & Standard dropout & SliceOut & speedups & memory gains \\
\midrule
CIFAR-10 & 28x6 & $96.1\%$ & $96.0\%$	& -25\%	& -21\% \\
& 34x8 & $96.2\%$ & $96.2\%$	& -26\%	& -21\% \\
& 40x10 & $96.3\%$ & $96.2\%$	& -30\%	& -20\% \\
& 46x12 & $96.4\%$ & $96.2\%$	& -29\%	& -25\% \\
& 52x14 & $96.4\%$ & $96.2\%$	& -32\%	& -22\% \\
\midrule
CIFAR-100 & 28x6 & $79.9\%$ & $79.8\%$	& -24\%	& -21\% \\
& 34x8  & $80.9\%$ & $80.7\%$	& -26\%	& -21\% \\
& 40x10 & $81.3\%$ & $81.2\%$	& -28\%	& -20\% \\
& 46x12 & $81.5\%$ & $81.4\%$	& -30\%	& -25\% \\
& 52x14 & $81.6\%$ & $81.5\%$	& -34\%	& -22\% \\
\bottomrule
\end{tabular}
}
\label{Table: WRN - Channel SliceOut & Probabilistic normalisation results}

\end{adjustbox}
\end{center}
\vspace{-4mm}
\end{table}

\subsection{EfficientNets}
\label{Experiment: EN}
EfficientNets \citep{tan2019efficientnet} achieve state of the art performance on several vision datasets including ImageNet \citep{russakovsky2014imagenet}, while being more compute efficient than prior architectures at test time. The purpose of our EfficientNets experiments is two-fold: first, we demonstrate the scalability and generalisability of the SliceOut scheme to larger datasets and more complex architectures; second, we show that SliceOut can also be thought of as a method to accelerate model training, even when dropout is not used in the original architecture. 
In EfficientNets, dropout is not used in any of the mobile inverted bottleneck (MBConv) blocks that form the backbone of the architecture. \footnote{Standard dropout is applied on the last fully connected layer of the network, but using SliceOut there would not result in meaningful speedups.} We use Channel-SliceOut to operate on the ``expand convolution'' (Fig.~\ref{EN - Architecture diagram}) of the first three stages of MBConv blocks, as this is where the largest tensors are created. Similar to what we observed with Wide ResNets, ``delayed normalisation'' (right before the final ``projection'' convolution) leads to higher test accuracy. We hypothesize that normalising earlier in the block leads to worse performance as it perturbs the statistics computed at train time for a given slice by the intermediate Batch normalisation layers.
Since we add SliceOut in parts of the network where no regularization is needed, we turn off SliceOut in the last $10\%$ of training epochs to bridge a potential gap in test accuracy with the original architecture (Appendix~\ref{Appendix: EfficientNets}).

In CIFAR-10/100 experiments, we fine-tune EfficientNet models pre-trained on ImageNet without SliceOut (following the same experimental setup as in \citet{tan2019efficientnet,kornblith2018better}). We observe speedups of over $20\%$ when using SliceOut, with comparable test accuracy to performing the fine tuning without SliceOut (Table.~\ref{Appendix - Table: EN results - CIFAR10/100}). This demonstrates that SliceOut can be used to achieve speedups when fine tuning networks that were pre-trained without. 
In ImageNet experiments, we train the networks from scratch and observe speedups of up to $20\%$ with SliceOut with similar test accuracy (Table~\ref{Table: EN results - ImageNet}), resulting in a more desirable compute efficiency frontier (Fig.~\ref{fig: Compute_efficacy frontiers}). We train a B2 architecture with SliceOut (0.4) as fast as a B1 architecture but with higher test accuracy (79.8\% Vs 78.8\%).

\begin{table}[!ht]
\begin{center}
\caption{\textbf{EfficientNet results - ImageNet.} Training time is the relative \% of train time per epoch for a network trained with SliceOut Vs standard B3 architecture trained on the same dataset without SliceOut. ImageNet results are obtained by training from scratch.}

\begin{adjustbox} {width=0.8\textwidth}
\footnotesize{
\begin{tabular}{c|cccc|cccc}
\toprule
SliceOut & \multicolumn{4}{c}{Test accuracy} & \multicolumn{4}{c}{Training time savings} \\
& & & & & \multicolumn{4}{c}{(rel. to B3 baseline)} \\
 rate & B0 & B1 & B2 & B3 & B0 & B1 & B2 & B3\\
\midrule
None &77.1\% &78.7\% &79.7\% &80.6\% &-68\% &-50\% &-39\% &0\% \\
0.3 &\textbf{77.2\%} &\textbf{78.8\%} &\textbf{79.8\%} &\textbf{81.0\%} &-70\% &-55\% &-46\% &-12\% \\
0.4 &76.8\% &\textbf{78.8\%} &\textbf{79.8\%} &80.7\% &-72\% &-57\% &-47\% &-18\% \\
0.5 &76.4\% &78.5\% &79.4\% &80.7\% &-72\% &-58\% &-52\% &-20\% \\
\bottomrule
\end{tabular}
}
\label{Table: EN results - ImageNet}

\end{adjustbox}
\end{center}
\vspace{-4mm}
\end{table}

\subsection{Transformers}
\label{Experiment: Transformers}

\begin{table}[h!]
\begin{center}
\caption{\textbf{Transformer results.} We observe speedups and memory gains of $\sim10\%$ when using SliceOut, despite the fact in Transformers the performance is dominated by looking up embedding vectors. Although Transformers are typically under-parametrised for language modeling on LM1B, SliceOut is a more effective form of regularization compared to standard dropout or controlled dropout (detailed results in Appendix~\ref{Appendix: Transformers}).}

\begin{adjustbox}{width=0.8\textwidth}
\footnotesize{
\begin{tabular}{cc|cc|ccc}
\toprule
Width & Dropout & Dropout & Controlled & SliceOut & Training & Max cached \\
 & rate & Perplexity & Perplexity & Perplexity & time & memory \\
\midrule
1024 & 0.0 & \textbf{31.7} & - & - & - & - \\
 & 0.3 & 45.1 & 45.7 & \textbf{33.7} & -8\% & -9\% \\
2048 & 0.0 & \textbf{28.1} & - & - & -& - \\
 & 0.3 & 88.6 & 53.1 & \textbf{28.1} & -11\% & -10\% \\
\bottomrule
\end{tabular}
}
\label{transformer}
\end{adjustbox}

\end{center}
\vspace{-2mm}
\end{table}

In our experiments, we evaluate a vanilla Transformer language model on the popular ``One Billion Word Benchmark'' \citep{chelba2013one}. Similar to what we do our EfficientNets experiments, we also perform a final fine tuning without SliceOut for the last 10\% epochs. Our results are shown in Table~\ref{transformer}. Given the complexity of language modeling on the LM1B dataset, we observe that test set perplexity is reduced across the board as we increase model width. In the larger width setting, we obtain identical perplexity with SliceOut as for models trained without (standard dropout is always detrimental to performance, see Appendix~\ref{Appendix: Transformers}), while reducing memory overhead by $\sim 10\%$ and reducing steptime by $\sim 10\%$.
Speedups are more modest in comparison to CNNs, and this is primarily due to the fact that, in Transformers, a significant portion of steptime is spent looking up embedding vectors and computing logits over a vocabulary of more than 32,000 elements. Similarly, much of the Transformer's memory is spent on storing the parameters, which SliceOut does not reduce. 

\section{Conclusion}

Training modern deep neural networks in a resource-intensive task. As deep learning-based applications become more pervasive, their impact on our environment is ever increasing \citep{alex2019quantifying-co2, strubell2019energy}. SliceOut is an effective approach for speeding up and reducing the memory requirements of neural networks at train time. This is particularly compelling when we need to frequently re-train the same models (e.g., active learning, continual learning) or train a large number of closely related architectures (e.g., hyperparameter search, neural architecture search). We demonstrated in this work how the scheme can be beneficial to a diverse set of network architectures across application domains.  Successfully leveraging SliceOut in large neural networks can help curtail the $CO_2$ emissions during training by 10-40\% (Appendix ~\ref{Appendix: CO2 emissions}).

\newpage
\small
\bibliography{references}

\newpage
\appendix
\section*{Appendix}
\section{SliceOut - Additional implementation details}

\subsection{Constraints on eligible positions for slicing}
\label{Appendix: Constraints on eligible positions for slicing}
\begin{figure*}[h]
    \centering
    \includegraphics[width=3in]{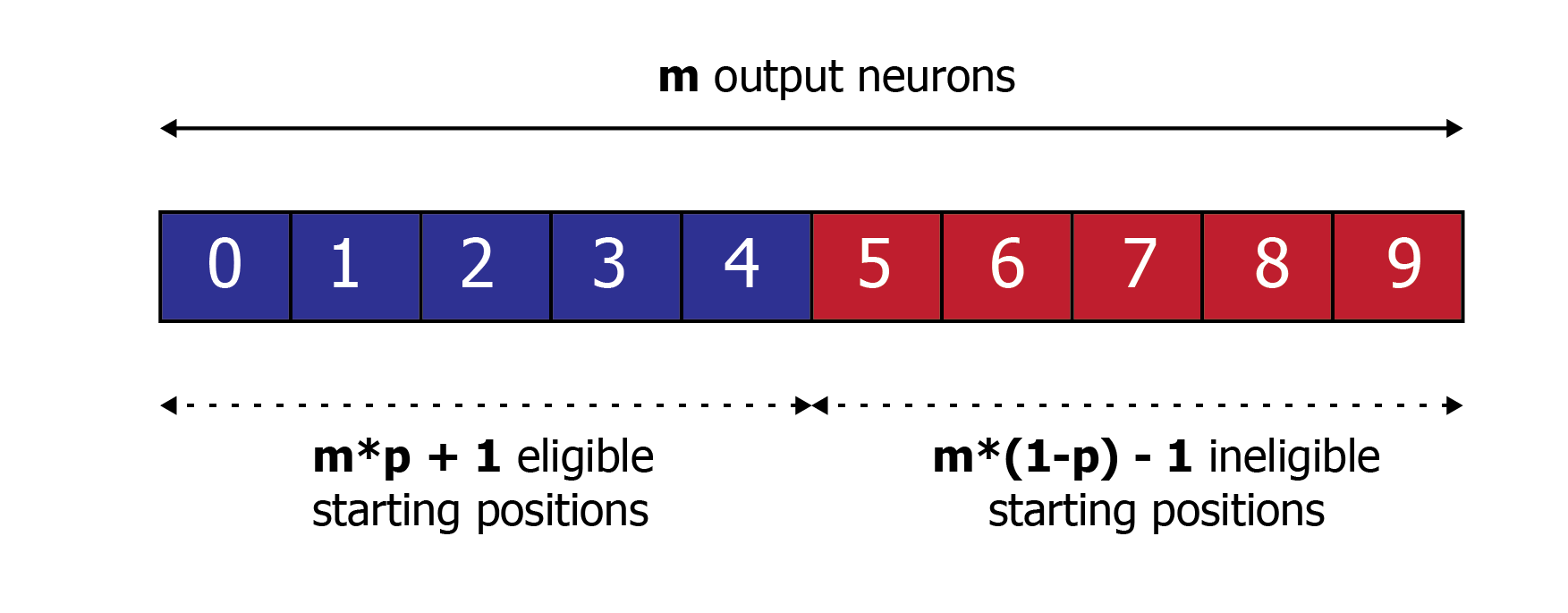}
    \caption{\textbf{Depiction of eligible indices for slice start -} In the example above, the layer width is 10, and dropout rate 40\%. Consequently, slices are contiguous sets of 6 units and the last 5 positions (i.e., indices 5 to 9) are ineligible to be the slice start to enforce strict same-sized slicing of contiguous units.}
    \label{Fig: Constraints on starting positions}
\end{figure*}

Modern GPU kernels have subroutines to select the best algorithm based on the shape of tensors involved in operations (e.g., cudnn.benchmark in Pytorch). These routines typically involve benchmarking different alternative algorithms in the first training step (e.g., FFT, Winograd), then keeping the best algorithm(s) for all subsequent steps, as long as the shape of tensors remains the same (otherwise the subroutine runs at each step). To obtain maximal speed-ups, we ensure that the shape of all tensors is constant throughout training, and consequently prevent re-running these optimisation subroutines. This is achieved by restricting the sampled starting index to a subset of eligible positions, specifically, restricting the starting index to be at a position where $start\_index + slide\_width \leq layer\_width $ (see Fig.~\ref{Fig: Constraints on starting positions} for a concrete example).

\section{Activation normalisation \& moment preservation}
\label{Appendix: Activation normalization & moment preservation}

We show here how the two normalisations schemes described in \S\ref{Section: SliceOut} -- exactly or approximately -- preserve the moments of the layer output (pre-activation). 

Let $x$ be the input tensor at a layer $W$, with $n$ input units \& $m$ output units, where the SliceOut function $\operatorname{S(.)}$ is applied. 
We note  $\operatorname{norm(.)}$ the normalisation operator used in SliceOut (``Flow'' or ``Probabilistic'' normalisation, as described in \ref{normalisation}), and $\mathbbm{1}\{.\}$ the Heaviside step function -- where $\mathbbm{1}\{unit\_j\ kept\}$ indicates whether unit $j$ in the layer is kept. $ \forall j \in \llbracket 1, m \rrbracket$, $\underline{e_j}$ is the canonical basis vector for the $j^{th}$ dimension of the output tensor space.

\subsection{First moment preservation}
\subsubsection{No SliceOut}
Using linearity of expectation, we obtain the following formula for the expected value of the output tensor:

\begin{equation}
\mathbb{E}(W.x) = \mathbb{E}(\sum_{j=1}^{m} (\sum_{i=1}^{n} w_{j,i} * x_i )\ \underline{e_j}) = \sum_{j=1}^{m} \sum_{i=1}^{n} w_{j,i} * \mathbb{E}(x_i)\ \underline{e_j} = W . \mathbb{E}(x)
\end{equation}

\subsubsection{With SliceOut}
Similarly, when SliceOut is applied we obtain the following since the expectation and normalisations operators are both linear and slicing is performed at random:

\begin{subequations}
\begin{align}
\mathbb{E}(\operatorname{S(W.x)}) & = \mathbb{E}(\operatorname{norm(\sum_{j=1}^{m} (\sum_{i=1}^{n} \mathbbm{1}\{unit\_j\ kept\} * \mathbbm{1}\{unit\_i\ kept\} * w_{j,i} * x_i )\ \underline{e_j})}) \\
& = \sum_{j=1}^{m} (\sum_{i=1}^{n} \mathbb{P}\{unit\_j\ kept\} * \mathbb{P}\{unit\_i\ kept\} * \operatorname{norm\_out(w_{j,i}) * \operatorname{norm\_in(\mathbb{E}(x_i))}})\ \underline{e_j} \\
& = \sum_{j=1}^{m} (\sum_{i=1}^{n} w_{j,i} * \mathbb{E}(x_i))\ \underline{e_j} = W . \mathbb{E}(x)
 \end{align}
\end{subequations}

In (2b) $\operatorname{norm\_out(.)}$ refers to the normalisation of the output, while $\operatorname{norm\_in(.)}$ refers to the normalisation of the input (relevant only if SliceOut was also applied on the prior layer).

Equality (2c) is exact when using the ``Probabilistic normalisation'', since the normalisation scheme consists exactly in dividing activations by the probability that the corresponding unit is kept. 
It is approximate in the case of the ``Flow'' normalisation, since we divide by the average probability that a unit is kept -- across the layer -- which differs from the probability that a given unit is kept (e.g., units near the edges are more likely to be dropped for reasons detailed in Appendix \ref{Appendix: Constraints on eligible positions for slicing}).

\subsection{Second moment preservation}

Let $y$ be the output tensor at layer $W$.

\subsubsection{No SliceOut}

\begin{equation}
\forall (j_1, j_2) \in \llbracket 1, m \rrbracket^2,  \operatorname{Cov}(y_{j_1}, y_{j_2})
 = \mathbb{E}(y_{j_1} * y_{j_2}) - \mathbb{E}(y_{j_1}) * \mathbb{E}(y_{j_2}) 
\end{equation}

\subsubsection{With SliceOut}

\begin{equation}
\forall (j_1, j_2) \in \llbracket 1, m \rrbracket^2,  \operatorname{Cov}(\operatorname{S(y_{j_1})},\operatorname{S(y_{j_2})}) = \mathbb{E}(\operatorname{S(y_{j_1})} * \operatorname{S(y_{j_2})}) - \mathbb{E}(\operatorname{S(y_{j_1}))} * \mathbb{E}(\operatorname{S(y_{j_2}))} 
\end{equation}

Since we have just seen preservation for the second term, we will focus on the first term only: 
\begin{subequations}
\begin{gather}
\mathbb{E}(\operatorname{S(y_{j_1})} * \operatorname{S(y_{j_2})})
 = \mathbb{E}(\operatorname{S(\sum_{i=1}^{n} w_{j_1,i} * x_i)} * \operatorname{S(\sum_{i=1}^{n} w_{j_2,i} * x_i)})
 \end{gather}
 
 \begin{multline}
 = \mathbb{E}(
 \sum_{i=1}^{n}\sum_{k=1}^{n} \mathbbm{1}\{unit\_j_1\ kept\} * \mathbbm{1}\{unit\_i\ kept\} * \mathbbm{1}\{unit\_j_2\ kept\} * \mathbbm{1}\{unit\_k\ kept\} * \\
 \operatorname{norm\_out(w_{j_1,i}) * \operatorname{norm\_in(x_i)}} * \operatorname{norm\_out(w_{j_2,k})} * \operatorname{norm\_in(x_k)}
 )
 \end{multline}

 \begin{multline}
 =
 \sum_{i=1}^{n}\sum_{k=1}^{n} \mathbbm{P}\{unit\_j_1\ kept\} *  \mathbbm{P}\{unit\_j_2\ kept| unit\_j_1\ kept\} * \mathbbm{P}\{unit\_i\ kept\} * \mathbbm{P}\{unit\_k\ kept| unit\_i\ kept\} * \\
 \operatorname{norm\_out(w_{j_1,i}) * \operatorname{norm\_in(x_i)}} * \operatorname{norm\_out(w_{j_2,k})} * \operatorname{norm\_in(x_k)}
 \end{multline}

\end{subequations}

When considering the ``Probabilistic normalisation'', we thus have:
\begin{equation}
\begin{split}
 \forall (j_1, j_2) \in \llbracket 1, m \rrbracket^2,\  & \mathbbm{E}(\operatorname{S(y_{j_1})},\operatorname{S(y_{j_2})}) = \sum_{i=1}^{n}\sum_{k=1}^{n} \mathbbm{P}\{unit\_j_2\ kept| unit\_j_1\ kept\}\ *\\ 
& \mathbbm{P}\{unit\_k\ kept| unit\_i\ kept\} * w_{j_1,i} * x_i * \operatorname{norm\_out(w_{j_2,k})} * \operatorname{norm\_in(x_k)}
\end{split}
\end{equation}

Unlike in standard dropout, for SliceOut we usually have:
\begin{equation}
\forall (j_1, j_2) \in \llbracket 1, m \rrbracket^2, \mathbbm{P}\{unit\_j_2\ kept| unit\_j_1\ kept\} \ne \mathbbm{P}\{unit\_j_2\ kept\}
\end{equation}
due to the structure imposed by slicing contiguous units, and therefore the second moment is not exactly preserved.

However, for dropout rates $p \leq 0.5$ we have: $\forall j_1 \in \llbracket slice\_width, m - slice\_width \rrbracket, \forall j_2 \in \llbracket 1, m \rrbracket,$
\begin{equation}
\begin{split}
 \mathbbm{P}\{unit\_j_1\ kept |unit\_j_2\ kept\} & = \mathbbm{P}\{unit\_j_1\ kept\} = 1 \\ 
\mathbbm{P}\{unit\_j_2\ kept |unit\_j_1\ kept\} & = \mathbbm{P}\{unit\_j_2\ kept\}.
\end{split}
\end{equation}

Thus we approximately have:
\begin{equation}
\forall (j_1, j_2) \in \llbracket 1, m \rrbracket^2,  \mathbbm{E}(\operatorname{S(y_{j_1})},\operatorname{S(y_{j_2})}) \sim \sum_{i=1}^{n}\sum_{k=1}^{n} w_{j_1,i} * x_i * w_{j_2,k} * x_k = \mathbbm{E}(\operatorname{y_{j_1}},\operatorname{y_{j_2}})
\end{equation}

Similarly, the ``Flow normalisation'' approximately preserves the second moment, with the same additional approximation discussed in the first moment section.

\section{Number of architectures sampled from}
\label{Appendix: Number of architectures sampled from}

\begin{figure*}[!ht]
    \centering
    \includegraphics[width=3in]{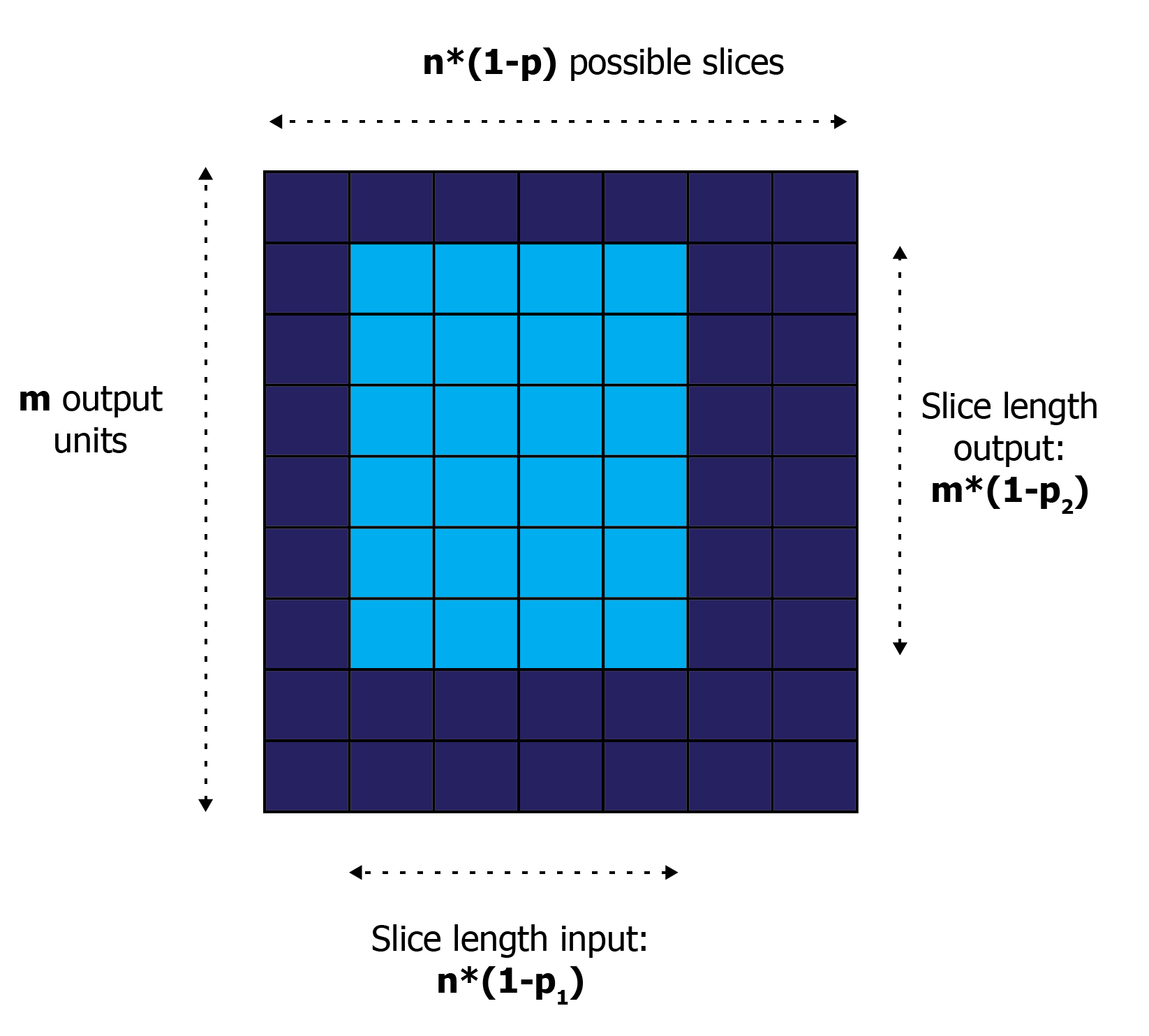}
    \caption{\textbf{Depiction of the SliceOut mask at a given fully-connected layer with n input units and m output units.}}
    \label{Fig: SliceOut mask}
\end{figure*}

Let's consider a simple feedforward network architecture with 3 hidden layers. We illustrate over a few scenarios the number of model architectures that SliceOut effectively samples from at \textbf{layer 2}, i.e., the number of distinct dropout masks at layer 2 resulting from applying SliceOut (see Fig.~\ref{Fig: SliceOut mask} for an example of a mask at layer 2). We assume that the first hidden layer has width n and the second hidden layer has width m. The SliceOut probabilities at these layers are respectively $p_1$ and $p_2$ (when SliceOut is indeed applied at the corresponding layer).

\subsection{Scenario 1: SliceOut applied only on hidden layer 1}
In that scenario, we apply SliceOut only on hidden layer 1, which means that we sample a contiguous set of $n*(1-p_1)$ output units from hidden layer 1. As a result, we also need to slice the columns of the weight matrix at layer 2 to match that same slice. There is a linear number of distinct slices of length $n*(1-p_1)$ we can take, therefore we sample from a linear number of architectures at layer 2.

\subsection{Scenario 2: SliceOut applied only on hidden layer 2}
This scenario is similar to scenario 1 except that we now need to slice the rows of the weight matrix at layer 2 to extract a slice of length $m*(1-p_2)$. There is also a linear number of distinct architectures sampled at layer 2.

\subsection{Scenario 3: SliceOut is applied on both hidden layers 1 and 2}
In that scenario, we both have to slice the columns and the rows of the weight matrix at layer 2. There are $n*(1-p_1) *m*(1-p_2) = O(n*m)$ distinct pairs of slices we can take on rows and columns of the weight matrix at layer 2, hence a quadratic number of distinct model architectures sampled from (quadratic in the layer width).

\section{Connection to BatchEnsemble} 
During training, the forward pass for a given layer where SliceOut is applied can be written as:
$y = f ((W(x \odot c )) \odot r )$
where W is the weight tensor, x is the activations tensor from the prior layer, y the activations tensor at this layer, and c and r are the dropout masks applied at this layer, operating respectively on the columns and rows of W.
This can be re-written: $y = f ((W\odot S)x)$, where $\odot$ is the Hadamard product between W and the rank-1 matrix $S=r \otimes c$.
\\
There is a striking parallel with the BatchEnsemble approach \citep{wen2020batchensemble}, a parameter-efficient approach to ensembling several models together, with two major differences: 
\begin{itemize}
    \item The parameters c and r are learned separately for each ensemble member in the BatchEnsemble case, while they are constant filters in SliceOut
    \item We are ensembling a sub-quadratic number of architectures in SliceOut, while there is no such restriction in the BatchEnsemble case (although the number of ensemble members is typically much smaller in practice, e.g., 2-10 models) 
\end{itemize}

\section{\texorpdfstring{CO$_2$ emissions calculation}{CO2 emissions calculation}}
\label{Appendix: CO2 emissions}

Given the speedups and memory gains it provides during training, SliceOut can help reduce the energy consumption, and thereby the $CO_2$ emissions, that would result from training a given model in three different ways:
\begin{itemize}
    \item \textbf{Approach 1 - Reduced number of machines needed during training:} Thanks to the memory efficiency provided by SliceOut, a model -- that would require a given number of GPUs when trained with standard dropout -- could be trained with fewer GPUs with SliceOut without degrading performance.
    For example, we saw in the Wide ResNets experiments on CIFAR-10 (Appendix~\ref{Appendix: WRN}) that a model with similar accuracy can be trained 39\% faster and with 23\% lower GPU memory requirements when using SliceOut (with a dropout rate 0.5). As a thought experiment, we compared the relative speed when training that model with SliceOut on 3 GPUs, compared to training the same model without SliceOut on 4 GPUs (keeping the ratio of CPUs per GPU constant): when training with SliceOut on 3 GPUs, the train time per epoch was 1\% lower. Therefore, training the same model on 25\% fewer GPU for about the same amount of time would result in $CO_2$ emissions that are 25\% lower;
    \item \textbf{Approach 2 - Same hardware, higher batch size:} Instead of reducing the number of GPUs to be used during training, we may want instead to further increase the batch size to make the most of the additional GPU memory available. 
    Using the same example as before, while this time keeping the same number of GPUs with and without SliceOut (1 GPU) but increasing the batch size so that the GPU memory footprint is identical with and without SliceOut, we observe new speedups from SliceOut -- and thus $CO_2$ emissions reductions -- of ~41\%;
    \item \textbf{Approach 3 - Same hardware, same hyperparameters, just faster training:} In cases where neither reducing the number of machines used during training nor increasing the batch size are desirable, training models with SliceOut can still help reduce $CO_2$ emissions through the speedups it provides, i.e., 10\% with Transformers or up to ~40\% in some of the CNNs architectures we analysed. 
\end{itemize}

To summarise, SliceOut can help train models with comparable accuracy while resulting in $CO_2$ emissions ~10\%-40\% lower, depending on the model architecture, the set of hyperparameters chosen and the hardware used for training.

\section{Detailed experimental results}
\label{Appendix: Experimental details}

In this section we present the comprehensive set of experimental results obtained across our experiments with fully-connected networks, Wide ResNets, EfficientNets and Transformers.

\subsection{Fully connected networks experiments}
\label{Appendix: Fully connected networks}

\subsubsection{Experimental setting}
\paragraph{Objectives.} In these experiments we aim to study the quality and stability of our SliceOut scheme in a simple setting and quantify the impact on memory savings and computation speedups over standard and controlled dropout. 
The purpose is not to obtain state-of-the-art performance on these datasets (the architecture leveraged here would not allow us to do that), but rather to confirm the validity and expected benefits of the SliceOut scheme with a simple architecture and commonly used datasets.

\paragraph{The MNIST and FashionMNIST datasets.} The MNIST \citep{Lecun1998MNIST} and FashionMNIST \citep{xiao2017fashionmnist} datasets consist of grayscale images of $28x28$ pixels, respectively representing the 10 digits and items from 10 distinct object classes from Zalando’s catalogue (e.g., shoes, bags, dresses). No data transformation is applied on the input.
FashionMNIST is available at the following location: https://github.com/zalandoresearch/fashion-mnist/tree/master/data/fashion.
MNIST is available at the following location: http://yann.lecun.com/exdb/mnist/.
We used the same train/test split as from these sources.

\paragraph{Model architecture.} For both the MNIST and FashionMNIST experiments we use the same model architecture: a simple fully connected network with 3-hidden layers. The input layer has 28x28 = 784 units (one for each image pixel), each hidden layer has 2048 units, and the output layer has 10 units (one for each class). We apply standard dropout, controlled dropout and SliceOut on each hidden layer, and vary the dropout rate from 0 to 0.9 by increment of 0.1.
For SliceOut, we compare the two types of normalisation discussed in \S~\ref{normalisation} -- in both cases, normalisation is applied immediately after SliceOut (unlike what we do in CNNs -- see Appendix~\ref{Appendix: CNNs}).

\paragraph{Training procedure and hyperparameters.} We minimize the cross-entropy loss between predictions and labels via the Adam algorithm. We used Pytorch (https://pytorch.org/docs/stable/index.html) to instantiate and train the different models. All hyperparameters are summarised in table \ref{Appendix - Table: FCN Hyperparameters}, using the Pytorch convention for the hyperparameter names (https://pytorch.org/docs/stable/optim.html).

\begin{table}[!ht]
\caption{MNIST \& FashionMNIST experiments - List of hyperparameters used for training}
\begin{center}
\begin{tabular}{cc}
\toprule
Hyperparameter & Value \\
\midrule
Batch size & 256 \\
Learning rate & $10^{-4}$\\
Beta1 & 0.9\\ 
Beta2 & 0.999 \\
Epsilon & $10^{-8}$\\
Weight decay & 0.0\\
\bottomrule
\end{tabular}
\end{center}
\label{Appendix - Table: FCN Hyperparameters}
\end{table}

\paragraph{Hardware.} All results for our FashionMNIST and MNIST experiments were obtained with a single GPU (Nvidia RTX 2080) and averaged across 4 independent runs.

\subsubsection{FashionMNIST experiments results}
\label{Appendix: FashionMNIST}
As discussed in \S~\ref{Experiment: FCN}, we observe speedups (up to 15\%), cached GPU memory savings (up to 30\%), faster convergence and to a higher test accuracy value with SliceOut, when typical dropout rates are applied (i.e., $p\leq 0.5$).
``Flow'' normalisation appears to be leading to higher test accuracy in this set of experiments compared to ``Probabilistic'' normalisation, although the difference is not always statistically significant (see Table~\ref{Appendix - Table: FashionMNIST test acc}).
We note that SliceOut may actually underperform compared to standard dropout, in the case of extreme dropout (e.g., $p \geq 0.9$) as it would lead to only ensembling under-capacitated networks.

Controlled dropout was more memory efficient than standard dropout only beyond a certain dropout rate ($p=0.4$), as the gains from storing smaller activations in memory (for the backward pass) become higher than the data duplication overhead resulting from the gather ops. SliceOut always consumed less memory as there is now such duplication.

\begin{table}[!ht]
\caption{FashionMNIST experiments - Highest test accuracy across dropout schemes}
\begin{center}
\begin{tabular}{c|cccc}
\toprule
Dropout rate & Standard dropout & SliceOut - & SliceOut - & Controlled dropout\\
& & Flow normalisation & Proba. normalisation & \\
\midrule
0.0 & $89.6\pm \SI{0.04}{\percent}$ & - & - & - \\
0.1 & $89.6\pm \SI{0.08}{\percent}$ & \textbf{90.0}$\pm \SI{0.02}{\percent}$ & \textbf{90.0}$\pm \SI{0.06}{\percent}$ & $89.8\pm \SI{0.05}{\percent}$ \\
0.2 & $89.4\pm \SI{0.03}{\percent}$ & \textbf{90.0}$\pm \SI{0.04}{\percent}$ & $89.7\pm \SI{0.03}{\percent}$ & $89.7\pm \SI{0.03}{\percent}$ \\
0.3 & $89.2\pm \SI{0.02}{\percent}$ & \textbf{90.0}$\pm \SI{0.07}{\percent}$ & $89.6\pm \SI{0.04}{\percent}$ & $89.6\pm \SI{0.05}{\percent}$ \\
0.4 & $89.0\pm \SI{0.05}{\percent}$ & $89.7\pm \SI{0.05}{\percent}$ & $89.5\pm \SI{0.05}{\percent}$ & $89.4\pm \SI{0.04}{\percent}$ \\
0.5 & $88.8\pm \SI{0.03}{\percent}$ & $89.6\pm \SI{0.05}{\percent}$ & $89.5\pm \SI{0.07}{\percent}$ & $89.3\pm \SI{0.04}{\percent}$ \\
0.6 & $88.5\pm \SI{0.03}{\percent}$ & $89.4\pm \SI{0.03}{\percent}$ & $89.2\pm \SI{0.05}{\percent}$ & $89.1\pm \SI{0.02}{\percent}$ \\
0.7 & $88.2\pm \SI{0.04}{\percent}$ & $89.1\pm \SI{0.02}{\percent}$ & $88.6\pm \SI{0.30}{\percent}$ & $88.7\pm \SI{0.02}{\percent}$ \\
0.8 & $87.8\pm \SI{0.01}{\percent}$ & $88.2\pm \SI{0.01}{\percent}$ & $87.7\pm \SI{0.03}{\percent}$ & $87.4\pm \SI{0.01}{\percent}$ \\
0.9 & $86.0\pm \SI{0.02}{\percent}$ & $72.9\pm \SI{0.17}{\percent}$ & $73.0\pm \SI{0.10}{\percent}$ & $66.8\pm \SI{1.87}{\percent}$ \\
\bottomrule
\end{tabular}
\end{center}
\label{Appendix - Table: FashionMNIST test acc}
\end{table}

\begin{table}[!ht]
\caption{FashionMNIST experiments - Train time per epoch (\% of standard dropout)}
\begin{center}
\begin{tabular}{c|ccc}
\toprule
Dropout rate & SliceOut - & SliceOut - & Controlled dropout\\
& Flow normalisation & Proba. normalisation & \\
\midrule
0.1 & $98.9\pm \SI{0.22}{\percent}$ & $98.5\pm \SI{0.76}{\percent}$ & $153.6\pm \SI{0.83}{\percent}$ \\
0.2 & $95.3\pm \SI{0.54}{\percent}$ & $95.4\pm \SI{0.68}{\percent}$ & $145.8\pm \SI{0.94}{\percent}$ \\
0.3 & $93.4\pm \SI{0.51}{\percent}$ & $93.4\pm \SI{0.54}{\percent}$ & $138.8\pm \SI{0.71}{\percent}$ \\
0.4 & $90.5\pm \SI{0.65}{\percent}$ & $91.2\pm \SI{0.84}{\percent}$ & $132.7\pm \SI{0.75}{\percent}$ \\
0.5 & $87.3\pm \SI{0.39}{\percent}$ & $87.8\pm \SI{0.60}{\percent}$ & $125.9\pm \SI{0.87}{\percent}$ \\
0.6 & $87.0\pm \SI{0.66}{\percent}$ & $87.5\pm \SI{0.63}{\percent}$ & $121.0\pm \SI{0.59}{\percent}$ \\
0.7 & $86.4\pm \SI{0.85}{\percent}$ & $86.9\pm \SI{0.76}{\percent}$ & $115.5\pm \SI{0.84}{\percent}$ \\
0.8 & $87.1\pm \SI{0.91}{\percent}$ & $86.7\pm \SI{0.80}{\percent}$ & $115.2\pm \SI{1.03}{\percent}$ \\
0.9 & $87.6\pm \SI{0.77}{\percent}$ & $86.9\pm \SI{0.83}{\percent}$ & $114.7\pm \SI{0.98}{\percent}$ \\
\bottomrule
\end{tabular}
\end{center}
\label{Appendix - Table: FashionMNIST train speed}
\end{table}

\begin{table}[!ht]
\caption{FashionMNIST experiments - Max GPU cached memory (\% of standard dropout). No confidence interval reported in the table below as the max memory usage was strictly equal across all experiments.}
\begin{center}
\begin{tabular}{c|ccc}
\toprule
Dropout rate & SliceOut - & SliceOut - & Controlled dropout\\
& Flow normalisation & Proba. normalisation & \\
\midrule
0.1 & 93.2\% & 93.2\% & 148.3\% \\
0.2 & 87.0\% & 87.0\% & 132.1\% \\
0.3 & 81.5\% & 81.5\% & 116.4\% \\
0.4 & 76.6\% & 76.6\% & 103.7\% \\
0.5 & 73.0\% & 73.0\% & 92.3\% \\
0.6 & 69.1\% & 69.1\% & 82.1\% \\
0.7 & 66.8\% & 66.8\% & 76.7\% \\
0.8 & 65.4\% & 65.4\% & 70.4\% \\
0.9 & 65.4\% & 65.4\% & 66.2\% \\
\bottomrule
\end{tabular}
\end{center}
\label{Appendix - Table: FashionMNIST max cached GPU memory}
\end{table}

\subsubsection{MNIST experiments results}
\label{Appendix: MNIST}
The results of the MNIST experiments are consistent with what we discussed in the FashionMNIST experiments in terms of speedups and memory gains. However, there was no statistically-significant difference in terms of test accuracy, all models convering to the same high test accuracy values ($\sim 98.5\%$).

\begin{table}[!ht]
\caption{MNIST experiments - Highest test accuracy across dropout schemes}
\begin{center}
\begin{tabular}{c|cccc}
\toprule
Dropout rate & Standard dropout & SliceOut - & SliceOut - & Controlled dropout\\
& & Flow normalisation & Proba. normalisation & \\
\midrule
0.1 & $98.4\pm \SI{0.03}{\percent}$ & $98.5\pm \SI{0.03}{\percent}$ & $98.5\pm \SI{0.04}{\percent}$ & $98.5\pm \SI{0.03}{\percent}$ \\
0.2 & $98.5\pm \SI{0.02}{\percent}$ & $98.5\pm \SI{0.03}{\percent}$ & $98.5\pm \SI{0.04}{\percent}$ & $98.6\pm \SI{0.02}{\percent}$ \\
0.3 & $98.5\pm \SI{0.04}{\percent}$ & $98.4\pm \SI{0.01}{\percent}$ & $98.4\pm \SI{0.04}{\percent}$ & $98.6\pm \SI{0.01}{\percent}$ \\
0.4 & $98.5\pm \SI{0.01}{\percent}$ & $98.5\pm \SI{0.03}{\percent}$ & $98.4\pm \SI{0.01}{\percent}$ & $98.6\pm \SI{0.02}{\percent}$ \\
0.5 & $98.6\pm \SI{0.03}{\percent}$ & $98.5\pm \SI{0.04}{\percent}$ & $98.4\pm \SI{0.05}{\percent}$ & $98.6\pm \SI{0.03}{\percent}$ \\
0.6 & $98.6\pm \SI{0.04}{\percent}$ & $98.5\pm \SI{0.03}{\percent}$ & $98.4\pm \SI{0.02}{\percent}$ & $98.6\pm \SI{0.02}{\percent}$ \\
0.7 & $98.6\pm \SI{0.02}{\percent}$ & $98.4\pm \SI{0.02}{\percent}$ & $98.3\pm \SI{0.02}{\percent}$ & $98.4\pm \SI{0.04}{\percent}$ \\
0.8 & $98.3\pm \SI{0.01}{\percent}$ & $98.0\pm \SI{0.02}{\percent}$ & $97.8\pm \SI{0.02}{\percent}$ & $97.6\pm \SI{0.01}{\percent}$ \\
0.9 & $97.3\pm \SI{0.02}{\percent}$ & $94.8\pm \SI{0.03}{\percent}$ & $94.4\pm \SI{0.07}{\percent}$ & $92.6\pm \SI{0.14}{\percent}$ \\
\bottomrule
\end{tabular}
\end{center}
\label{Appendix - Table: MNIST test acc}
\end{table}

\begin{table}[!ht]
\caption{MNIST experiments - Train time per epoch (\% of standard dropout)}
\begin{center}
\begin{tabular}{c|ccc}
\toprule
Dropout rate & SliceOut - & SliceOut - & Controlled dropout\\
& Flow normalisation & Proba. normalisation & \\
\midrule
0.1 & $98.3\pm \SI{0.38}{\percent}$ & $97.6\pm \SI{0.69}{\percent}$ & $153.9\pm \SI{0.85}{\percent}$ \\
0.2 & $94.7\pm \SI{0.52}{\percent}$ & $95.4\pm \SI{0.70}{\percent}$ & $144.7\pm \SI{1.09}{\percent}$ \\
0.3 & $93.1\pm \SI{0.79}{\percent}$ & $92.8\pm \SI{0.83}{\percent}$ & $138.1\pm \SI{0.71}{\percent}$ \\
0.4 & $91.0\pm \SI{0.74}{\percent}$ & $90.9\pm \SI{0.93}{\percent}$ & $131.4\pm \SI{1.07}{\percent}$ \\
0.5 & $88.5\pm \SI{0.70}{\percent}$ & $88.6\pm \SI{0.71}{\percent}$ & $126.8\pm \SI{0.65}{\percent}$ \\
0.6 & $86.7\pm \SI{0.51}{\percent}$ & $87.4\pm \SI{0.98}{\percent}$ & $119.9\pm \SI{0.58}{\percent}$ \\
0.7 & $87.6\pm \SI{1.20}{\percent}$ & $86.4\pm \SI{0.60}{\percent}$ & $116.0\pm \SI{0.66}{\percent}$ \\
0.8 & $88.1\pm \SI{0.86}{\percent}$ & $86.0\pm \SI{0.76}{\percent}$ & $115.6\pm \SI{0.68}{\percent}$ \\
0.9 & $86.0\pm \SI{0.55}{\percent}$ & $84.6\pm \SI{0.47}{\percent}$ & $113.5\pm \SI{0.80}{\percent}$ \\
\bottomrule
\end{tabular}
\end{center}
\label{Appendix - Table: MNIST train speed}
\end{table}

\begin{table}[!ht]
\caption{MNIST experiments - Max GPU cached memory (\% of standard dropout). No confidence interval reported in the table below as the max memory usage was strictly equal across all experiments.}
\begin{center}
\begin{tabular}{c|ccc}
\toprule
Dropout rate & SliceOut - & SliceOut - & Controlled dropout\\
& Flow normalisation & Proba. normalisation & \\
\midrule
0.1 & 93.2\% & 93.2\% & 148.4\% \\
0.2 & 87.0\% & 87.0\% & 132.0\% \\
0.3 & 81.5\% & 81.5\% & 116.2\% \\
0.4 & 76.6\% & 76.6\% & 104.5\% \\
0.5 & 73.0\% & 73.0\% & 92.3\% \\
0.6 & 69.1\% & 69.1\% & 82.3\% \\
0.7 & 66.8\% & 66.8\% & 76.5\% \\
0.8 & 65.4\% & 65.4\% & 70.5\% \\
0.9 & 65.4\% & 65.4\% & 66.2\% \\
\bottomrule
\end{tabular}
\end{center}
\label{Appendix - Table: MNIST max cached GPU memory}
\end{table}

\subsection{Convolutional neural networks experiments}
\label{Appendix: CNNs}

\subsubsection{Wide ResNets}
\label{Appendix: WRN}

\paragraph{Objectives.} The main objective of our Wide ResNets experiments was to demonstrate that the benefits of the method are even greater with convolutional neural networks due to the relatively higher footprint of activations in GPU memory (because of weight sharing), and to show that these benefits can be achieved on a larger and more complex set of data.
We chose to analyse the benefits of SliceOut in Wide ResNets \citep{zagoruyko2016wide} as they achieve high test accuracy across many vision tasks, and include dropout in each residual block as a way to mitigate potential overfitting risk due to the channel widening.

\paragraph{The CIFAR-10 and CIFAR-100 datasets.} The CIFAR-10 and CIFAR-100 datasets contain 32x32 color images of respectively 10 and 100 distinct object or animal classes (e.g., airplane, automobile, cat, dog). 
The datasets can be obtained at the following location: https://www.cs.toronto.edu/~kriz/cifar.html. 
We used the same train / test split as from the source website, i.e., 50k images in the training data and 10k images in the test data.
Following the data preparation approach of the original Wide ResNet paper \citep{zagoruyko2016wide}, our data augmentation consisted only of random horizontal flips and random crops of 224 pixels.

\paragraph{Model architecture.} The overall model architecture used in our experiments follows very closely the original Wide ResNets architecture. The only notable difference is the type of dropout used across experiments (standard dropout Vs SliceOut) -- see detailed architecture diagram in Fig.\ref{WRN - Architecture diagram}.
When using SliceOut, it was critical to align the sampled slices between the two convolution layers and the batch norm in-between (all items in orange font on Fig.\ref{WRN - Architecture diagram}).

We experimented with the ``Flow'' and ``Probabilistic'' normalisations schemes, and observed systematically higher test accuracy with the latter. ``Delaying'' the normalisation until after the batch norm and right before the second convolution also helped further increase accuracy.

We additionally compared the performance of the two types of SliceOut described in \S~\ref{SliceOut and CNNs}, and observed higher test accuracy with ``Channel SliceOut'' over ``Patch SliceOut'' (see Tables \ref{Appendix - Table: WRN - Channel SliceOut & Probabilistic normalisation results} and \ref{Appendix - Table: WRN results - Patch Flow}).

\begin{figure*}
    \centering
    \includegraphics{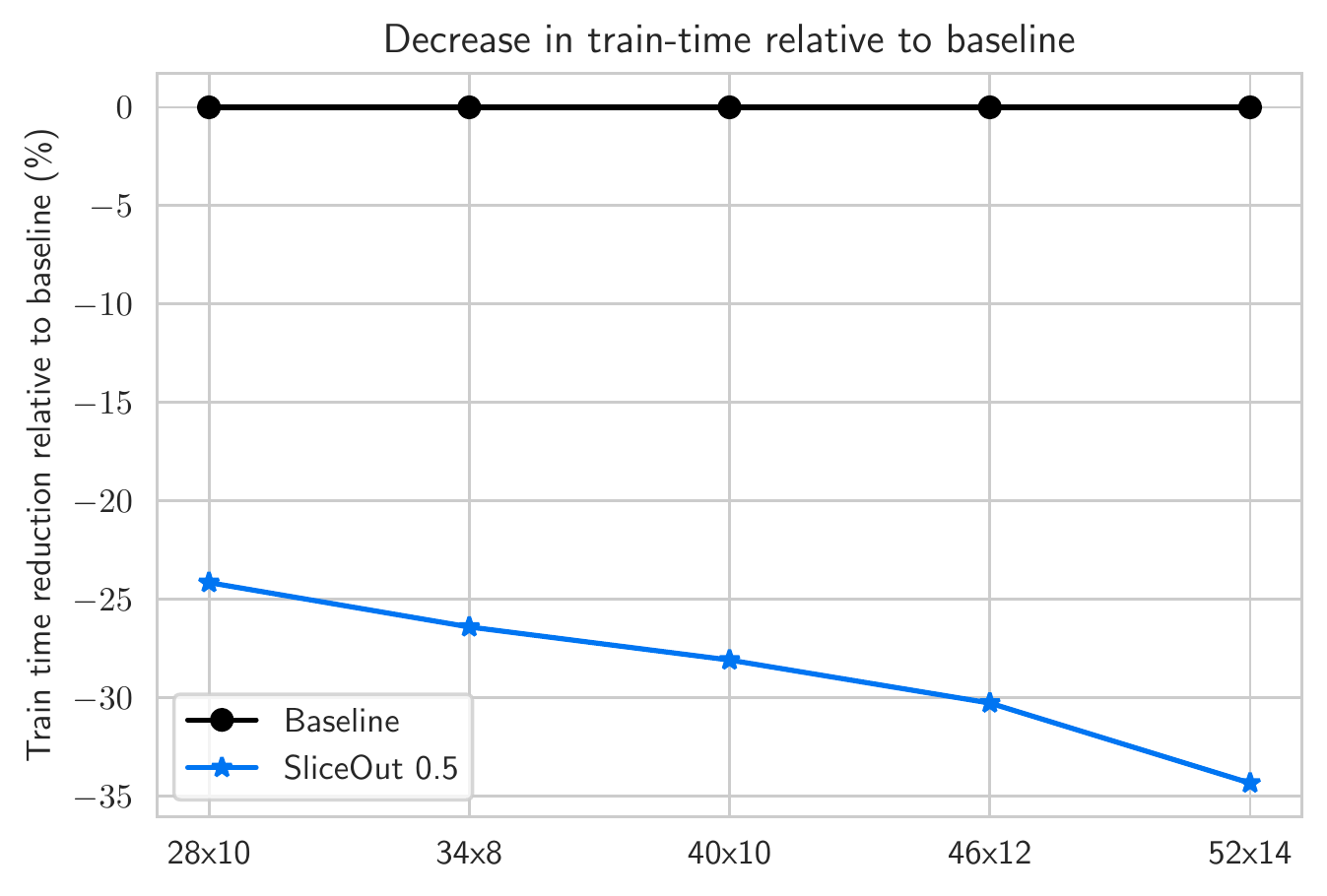}
    \caption{\textbf{Training speedups in Wide ResNets on CIFAR100} For a given Wide ResNet architecture, we can achieve speedups of up to 35\% with SliceOut (Channel SliceOut with probabilistic normalisation and SliceOut rate of 0.5), with no impact on test accuracy (see also Table~\ref{Table: WRN - Channel SliceOut & Probabilistic normalisation results}).}
    \label{Fig_Appendix_WRN_speedups_by_architecture}
\end{figure*}

We investigated the benefits of SliceOut on the 40x10 Wide ResNet architecture since authors of the original Wide ResNet paper reported top test accuracy on CIFAR-10 and CIFAR-100 with this architecture. We then conducted additional experiments with smaller (28x6 and 34x8) and larger (46x12 and 52x14) architectures to understand how training speedups would be impacted as we scale up or down (see Fig.~\ref{Fig_Appendix_WRN_speedups_by_architecture}).

\paragraph{Training procedure and hyperparameters.} We closely followed the training process from the original Wide ResNet paper \citep{zagoruyko2016wide}, i.e. minimising the cross-entropy loss via SGD with momentum over 200 epochs, with an identical learning rate schedule. We summarise all the hyperparameters used in the table below:

\begin{table}[!ht]
\caption{Wide ResNets experiments - List of hyperparameters used for training}
\begin{center}
\begin{tabular}{cc}
\toprule
Hyperparameter & Value \\
\midrule
Batch size & 128 \\
Initial learning rate & 0.1 \\
Learning rate schedule & Dropped by 0.2 at epochs 60,120 and 160\\
Momentum & 0.9\\ 
Dampening & 0.0 \\
Weight decay & $5*10^{-4}$\\
\bottomrule
\end{tabular}
\end{center}
\label{Appendix - Table: WRN Hyperparameters}
\end{table}

\paragraph{Hardware.} All our Wide ResNets results were obtained by running experiments (five independent runs) on a single GPU (Nvidia Titan RTX).

\paragraph{Experiments results.} We observe speedups of up to $\sim 40\%$ and memory gains of up to $\sim 30\%$ across runs, with a test accuracy that matches the value obtained for the best baseline Wide ResNet models (with standard dropout).
In our best Wide ResNets results with ``Channel Sliceout'' and ``Probabilistic normalisation'', we obtain increased performance (higher final test accuracy Vs training speedups) when turning off SliceOut in the first residual block of the network and the last block of each group (i.e., 4 residual blocks in total).
``Channel SliceOut'' consistently outperformed ``Patch SliceOut'' across experiments. The ``Probabilistic'' normalisation delivered higher test accuracy over the ``Flow'' normalisation for ``Channel SliceOut'', while ``Patch SliceOut'' had better results with the ``Flow'' normalisation (Tables \ref{Appendix - Table: WRN - Channel SliceOut & Probabilistic normalisation results} and \ref{Appendix - Table: WRN results - Patch Flow}).

\begin{table}[!ht]
\begin{center}
\caption{\textbf{Wide ResNets - Channel SliceOut with ``Probabilistic'' normalisation} Training time \& Max cached GPU memory are respectively the relative \% of train time per epoch for a network trained with SliceOut Vs standard dropout, and the maximum cached GPU memory during training. SliceOut is turned off in the first residual block of the network and the last block of each group. Results are obtained with a $40\text{x}10$ architecture, and averaged over 5 independent runs. }

\begin{adjustbox}{width=0.9\textwidth}
\footnotesize{
\begin{tabular}{cc|c|ccc}
\toprule
Dataset & Architecture & Test accuracy & Test accuracy & Training & Max cached \\
 &  rate & Standard dropout & SliceOut & time & memory \\
\midrule
 CIFAR-10 & 0.0 & $\textbf{96.3\%}$ & - & - & - \\
 & 0.1 & $\textbf{96.3\%}$ & $\textbf{96.3\%}$ & -5\% & -8\% \\
 & 0.2 & $96.0\%$ & $\textbf{96.2\%}$ & -17\% & -10\% \\
 & 0.3 & $95.6 \%$ & $\textbf{96.2\%}$ & -21\% & -13\% \\
 & 0.4 & $94.7 \%$ & $\textbf{96.2\%}$ & -26\% & -15\% \\
 & 0.5 & $93.4\%$ & $\textbf{96.2\%}$ & -30\% & -20\% \\
\midrule
CIFAR-100 & 0.0 & $81.2\%$ & - & - & - \\
 & 0.1 & $\textbf{81.3\%}$ & $\textbf{81.4\%}$ & -5\% & -8\% \\
 & 0.2 & $80.6\%$ & $\textbf{81.3\%}$ & -17\% & -10\% \\
 & 0.3 & $78.2\%$ & $81.2\%$ & -20\% & -13\% \\
 & 0.4 & $75.3\%$ & $81.2\%$ & -26\% & -15\% \\
 & 0.5 & $74.1\%$ & $81.2\%$ & -28\% & -20\% \\
\bottomrule
\end{tabular}
}
\label{Appendix - Table: WRN - Channel SliceOut & Probabilistic normalisation results}

\end{adjustbox}
\end{center}
\vspace{-4mm}
\end{table}

\begin{table}[!ht]
\begin{center}
\caption{\textbf{Wide ResNets - Patch SliceOut with ``Flow'' normalisation} Training time \& Max cached GPU memory are resp. the relative \% of train time per epoch for a network trained with SliceOut Vs standard dropout, and the maximum cached GPU memory during training. SliceOut applied uniformly to all residual blocks in the architecture. Results are obtained with a $40\text{x}10$ architecture, and averaged over 5 independent runs.}

\footnotesize{
\begin{tabular}{ccc|c|ccc}
\toprule
Dataset & Architecture & Dropout & Test accuracy & Test accuracy & Training & Max cached \\
 &  & rate & Standard dropout & SliceOut & time & memory \\
\midrule
CIFAR-10 & 40x10 & 0.0 & \textbf{96.3\%} & - & - & - \\
 &  & 0.1 & \textbf{96.3\%} & 96.0\% & -5\% & -8\% \\
 &  & 0.2 & 96.0\% & 95.8\% & -7\% & -13\% \\
 &  & 0.3 & 95.6\% & 95.0\% & -10\% & -18\% \\
 &  & 0.4 & 94.7\% & 94.9\% & -23\% & -23\% \\
 &  & 0.5 & 93.4\% & 92.5\% & -26\% & -24\% \\
\midrule
CIFAR-100 & 40x10 & 0.0 & \textbf{81.2\%} & - & - & - \\
 &  & 0.1 & \textbf{81.3\%} & 80.5\% & -5\% & -8\% \\
 &  & 0.2 & 80.6\% & 80.0\% & -7\% & -13\% \\
 &  & 0.3 & 78.2\% & 78.4\% & -11\% & -18\% \\
 &  & 0.4 & 75.3\% & 76.9\% & -24\% & -23\% \\
 &  & 0.5 & 74.1\% & 73.6\% & -26\% & -24\% \\
 \bottomrule
\end{tabular}
}
\label{Appendix - Table: WRN results - Patch Flow}
\end{center}
\end{table}

\subsubsection{EfficientNets}
\label{Appendix: EfficientNets}

\paragraph{Objectives.} The objectives of the EfficientNets experiments were as follows:
\begin{itemize}
    \item  Demonstrate the scalability and generalisability of the SliceOut scheme to larger datasets and more complex architectures;
    \item Illustrate that SliceOut can be used more broadly as a tool to achieve speedups, even when the original architecture did not use (standard) dropout in the first place;
    \item Show that SliceOut can be used effectively to fine-tune a model that was originally trained without it.
\end{itemize}

\paragraph{Datasets.} We leveraged the CIFAR-10/100 datasets, as described in \S~\ref{Appendix: WRN}, and the ImageNet 2012 classification dataset (http://image-net.org/challenges/LSVRC/2012/index), comprised of over 1 million training images across 1,000 distinct classes \citep{russakovsky2014imagenet}.
Regarding data augmentation, we strictly follow the choices from the original EfficientNet paper \citep{tan2019efficientnet}:
\begin{itemize}
    \item CIFAR10/100 or ImageNet auto-augment policies
    \item Random horizontal flips
    \item Bicubic image interpolation to resize images to the required resolution needed by the different EfficientNet architectures
\end{itemize}

\paragraph{Model architecture.} Standard dropout is not used in any of the mobile inverted bottleneck (MBConv) blocks forming the backbone of the EfficientNet architecture (standard dropout is only applied on the last fully connected layer, offering very limited opportunity to obtain speedups and memory gains if replaced by SliceOut).
We apply SliceOut on the very first convolution layer in each MBConv block (see Fig.~\ref{EN - Architecture diagram}) -- this is where the channel width expansion is being performed, and typically where the largest tensors are being created. We propagate the sliced tensor throughout the block, up until the final "Project block" where we perform the ``delayed normalisation'' (see following paragraph).
Similar to what we discussed with Wide ResNets, it is critical to ensure the same slice is being used throughout the block (all elements in orange font in Fig.~\ref{EN - Architecture diagram}).
We also limit the usage of SliceOut to the first three ``stages'' (as per the terminology of the original paper) of MBConv layers -- they represent the highest opportunity for SliceOut as the largest tensors are created there.

\paragraph{Delayed normalisation.} One of the key characteristics of the MBConv block used in EfficienNets is the fact that, after the "expand convolution" is performed, each channel is dealt with independently from other channels by the subsequent two blocks (depthwise and squeeze-and-excite blocks). We found experimentally that delaying the (probabilistic) normalisation of activations after the squeeze-and-excite layer, and right before the projection convolution, provided the best stability to learning and ultimately the highest test accuracy. As discussed in \S~\ref{Section: Experimental results}, we hypothesize that normalising earlier in the block leads to worse performance as it perturbs the statistics computed at train time for a given slice by the intermediate Batch normalisation layers.

\paragraph{Training procedure and hyperparameters.} 
For the ImageNet experiments, we train networks from scratch using the architecture and hyperparameters detailed in the original EfficientNet paper \citep{tan2019efficientnet}. We used a batch size of 512, as larger batch sizes did not fit the 4-GPU machines we used for training. Following \citet{goyal2017largeminibatchImagenet}, we linearly scale the learning rate accordingly. During the last 10\% of training (i.e., last 35 epochs), we turn off SliceOut and ``fine-tune'' the full architecture with a constant learning rate of $10^{-4}$.
 For the CIFAR 10/100 experiments, we fine-tune EfficientNet models pre-trained on ImageNet (without SliceOut), with SGD over 200 epochs.
 Hyper-parameters chosen for all experiments are summarized in Table~\ref{Appendix - Table: EN Hyperparameters - Imagenet} and Table~\ref{Appendix - Table: EN Hyperparameters - CIFAR 10/100}. 

\begin{table}[!ht]
\caption{EfficientNet ImageNet experiments - List of hyperparameters used for training}
\begin{center}
\begin{tabular}{cc}
\toprule
Hyperparameter & Value \\
\midrule
Batch size & 512 \\
Number of epochs & 350 (last 35 epochs turning off SliceOut) \\
Optimizer type & RMSprop \\
Initial learning rate & 0.032 \\
Learning rate schedule & Exponential decay by 0.97 every 2.4 epochs\\
Momentum & 0.9\\ 
Dampening & 0.9 \\
Weight decay & $10^{-5}$\\
Batch norm momentum & 0.99\\
\bottomrule
\end{tabular}
\end{center}
\label{Appendix - Table: EN Hyperparameters - Imagenet}
\end{table}

\begin{table}[!ht]
\caption{EfficientNet CIFAR 10/100 experiments - List of hyperparameters used for training}
\begin{center}
\begin{tabular}{cc}
\toprule
Hyperparameter & Value \\
\midrule
Batch size & 128 \\
Number of epochs & 200 (last 20 epochs turning off SliceOut) \\
Optimizer type & SGD \\
Initial learning rate & 0.02 \\
Learning rate schedule & Exponential decay by 0.985 every epoch\\
Momentum & 0.9\\ 
Weight decay & $10^{-5}$\\
Batch norm momentum & 0.99\\
\bottomrule
\end{tabular}
\end{center}
\label{Appendix - Table: EN Hyperparameters - CIFAR 10/100}
\end{table}

\paragraph{Hardware.} We conducted all our experiments on single compute nodes with 4 Titan RTX GPUs.

\paragraph{Experiments results.}
We observe speedups of up to ~20\% with SliceOut, with similar test accuracy across the different experiments (Table~\ref{Table: EN results - ImageNet} and Table~\ref{Appendix - Table: EN results - CIFAR10/100}).

\begin{figure*}
    \centering
    \includegraphics{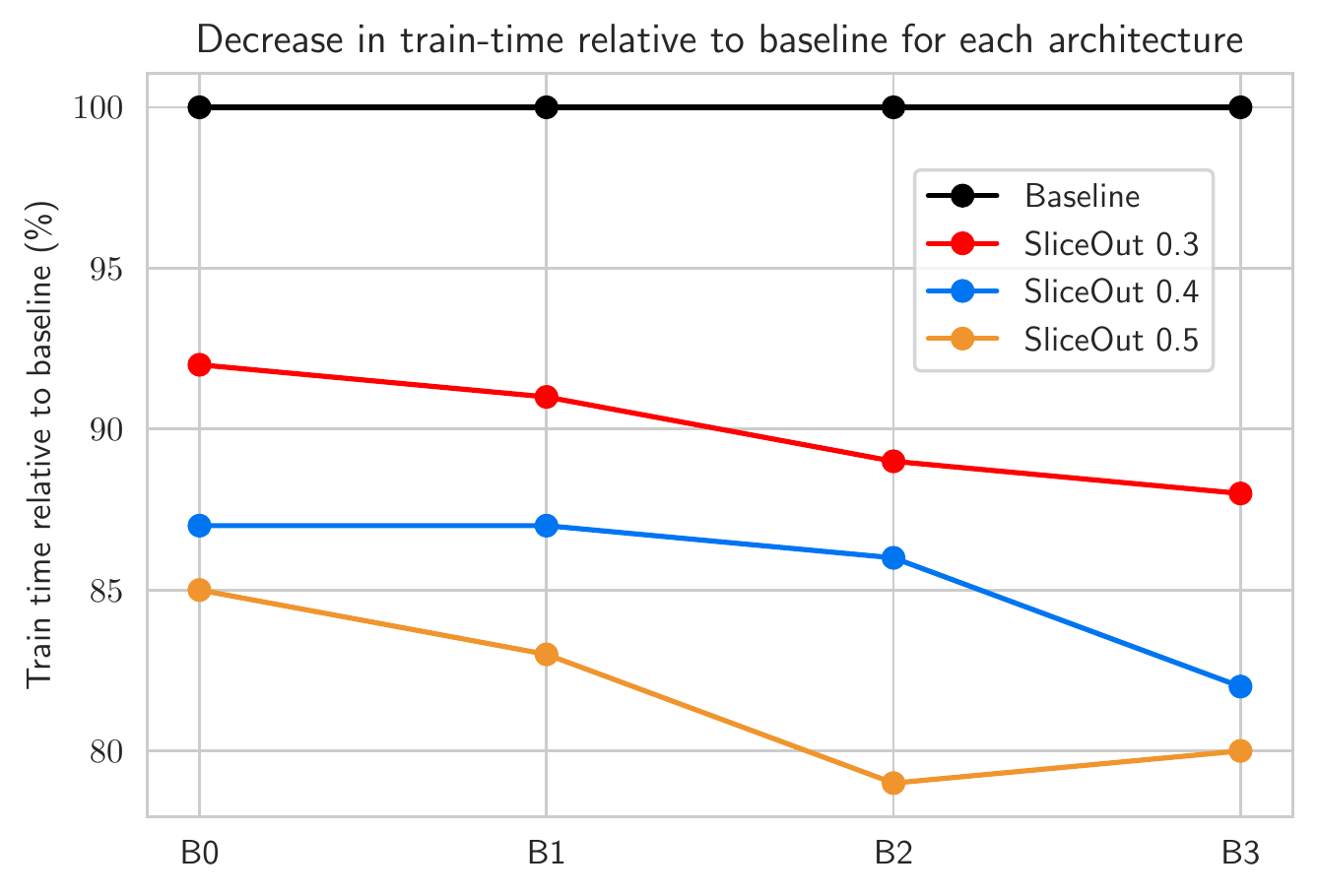}
    \caption{\textbf{Training speedups in EfficientNets on Imagenet} For a given EfficientNet architecture, we can achieve speedups of up to 20\%, with no impact on test accuracy (see also Table~\ref{Table: EN results - ImageNet}).}
\end{figure*}

\begin{table}[!ht]
\begin{center}
\caption{\textbf{EfficientNet CIFAR-10/100 results.} Training time is the relative \% of train time per epoch for a network trained with SliceOut Vs standard B2 architecture trained on the same dataset without SliceOut. CIFAR-10/100 results are obtained by fine tuning models pre-trained on ImageNet (pre-training without SliceOut).}

\begin{adjustbox} {width=0.7\textwidth}
\footnotesize{
\begin{tabular}{cc|ccc|ccc}
\toprule
Dataset & SliceOut & \multicolumn{3}{c}{Test accuracy} & \multicolumn{3}{c}{Training time} \\
 & rate & B0 & B1 & B2 & B0 & B1 & B2 \\
\midrule
 CIFAR-10 & None &98.3\% &98.4\% &98.5\% &-42\% &-16\% &0\% \\
 & 0.3 &98.1\% &98.4\% &98.4\% &-46\% &-24\% &-10\% \\
 & 0.4 &98.0\% &98.2\% &98.3\% &-47\% &-25\% &-13\% \\
 & 0.5 &97.9\% &98.1\% &98.3\% &-48\% &-27\% &-16\% \\
 \midrule
 CIFAR-100 & None &88.2\% &88.5\% &89.0\% &-42\% &-15\% &0\% \\
 & 0.3 &87.7\% &87.9\% &88.6\% &-45\% &-23\% &-11\% \\
 & 0.4 &87.3\% &87.8\% &88.6\% &-47\% &-24\% &-13\% \\
 & 0.5 &86.6\% &87.7\% &88.0\% &-48\% &-27\% &-17\% \\
\bottomrule
\end{tabular}
}
\label{Appendix - Table: EN results - CIFAR10/100}

\end{adjustbox}
\end{center}
\vspace{-4mm}
\end{table}

\subsection{Transformers}
\label{Appendix: Transformers}
\paragraph{Objectives.}
The purpose of our Transformers experiments was to demonstrate the benefits of SliceOut in a different application domain (language modelling), with another model architecture and with a larger dataset.

\paragraph{The LM1B Dataset.} 
The L1MB dataset is a benchmark corpus for measuring progress in statistical language modeling, with about one billion words in the training data. The dataset can be obtained at the following location: http://www.statmt.org/lm-benchmark/. 
We used the same train / test split as from the source website, with 30,301,028 examples in training data and 306,688 examples in test data.

\paragraph{Model architecture.}
The model architecture follows very closely the original Transformer architecture \citep{vaswani2017attention}. Each model had 6 encoder and decoder layers and the number of attention heads was set to 4. The dimension of feed-forward network was set to 4 times the embedding dimension. We experimented with the normalisation schemes discussed in \S~\ref{normalisation} and observed that the ``Probabilistic'' normalisation performed best.

\paragraph{Training procedure and hyperparameters.}
We followed the same training process as in the original Transformer paper \citep{vaswani2017attention} (e.g., stochastic optimisation with Adam algorithm, same learning rate schedule as in the original paper). 
We trained the different models with batch sizes of 256 and 1024, and selected the value providing the lowest perplexity for each model. 
The models with embedding dimension of 1024 were trained for 240,000 steps and the model with embedding dimension of 2048 were trained for 300,000 steps. The models trained with SliceOut were finetuned without SliceOut for the last 1\% of the training steps.

\paragraph{Hardware.}
Models with embedding dimension of 1024 were trained on 64-core TPUv3 clusters and 2048 or higher were trained on 256-core TPUv3 clusters.

\paragraph{Experiment results.} Transformers are typically underparametrised for a dataset as rich and complex as LM1B. Consequently, any type of regularisation on the ``smaller'' Transformer models we trained (with embedding size 1024) leads to higher perplexity than a model trained with no dropout. We note however that SliceOut outperforms the other types of regularisation we experimented with (standard dropout and controlled dropout) for all dropout rates tested (0.1 to 0.3).
As we increase the size of the embedding from 1024 to 2048, the perplexity of all models decreases significantly (down to $\sim 28$). However, the gap in perplexity between the models trained with SliceOut and without is negligible ($\sim 0.05$ perplexity points), and we are able to preserve the same speedups and memory gains ($\sim 10\%$).
\begin{table}[!ht]
\begin{center}
\caption{Transformer results}
\footnotesize{
\begin{tabular}{cc|cc|ccc}
\toprule
Width & Dropout & Dropout & Controlled & SliceOut & Training & Max cached \\
 & rate & Perplexity & Perplexity & Perplexity & time & memory \\
\midrule
1024 & 0.0 & 31.7 & - & - & - & - \\
 & 0.1 & 33.0 & 33.0 & \textbf{32.4} & 6\% & -1\% \\
 & 0.2 & 43.5 & 36.4 & \textbf{32.9} & -3\% & -5\% \\
 & 0.3 & 45.1 & 45.7 & \textbf{33.7} & -8\% & -9\% \\
2048 & 0.0 & 28.1 & - & - & -& - \\
 & 0.3 & 88.6 & 53.1 & \textbf{28.1} & -11\% & -10\% \\
\bottomrule
\end{tabular}
}
\label{transformer_detailed_table}
\end{center}
\end{table}

\pgfplotsset{width=12.5cm,height=8cm,compat=1.9}

\begin{center}
\begin{tikzpicture}
\begin{axis}[
    title={Perplexity vs No. of steps: 1024 embedding dim.},
    xlabel={No. of steps (in 1000s)},
    xmin=0, xmax=240,
    ymin=30, ymax=70,
    xtick={0,40,80,120,160,200, 240},
    ytick={30,40,50,60,70},
    legend pos=north east,
    ymajorgrids=true,
    grid style=dashed,
    title style={font=\small},
    label style={font=\small},
    legend style={font=\tiny},
]
\addplot[color=black,]
  coordinates {
(7.68,61.720)	(15.36,48.639)	(23.04,43.915)	(30.72,41.425)	(38.4,39.691)	(46.08,38.481)	(53.76,37.513)	(61.44,36.738)	(69.12,36.150)	(76.8,35.586)	(84.48,35.199)	(92.16,34.815)	(99.84,34.443)	(107.52,34.161)	(115.2,33.861)	(122.88,33.661)	(130.56,33.444)	(138.24,33.323)	(145.92,33.145)	(153.6,32.958)	(161.28,32.834)	(168.96,32.666)	(176.64,32.582)	(184.32,32.385)	(192.0,32.294)	(199.68,32.194)	(207.36,32.076)	(215.04,31.945)	(222.72,31.845)	(230.4,31.736)	(238.08,31.671)	(240.0,31.673)
  };
\addplot[color=brown,]
    coordinates {
(7.5,64.172)	(15.0,50.478)	(22.5,45.713)	(30.0,42.999)	(37.5,41.178)	(45.0,39.943)	(52.5,38.949)	(60.0,38.258)	(67.5,37.639)	(75.0,37.090)	(82.5,36.649)	(90.0,36.297)	(97.5,35.893)	(105.0,35.558)	(112.5,35.239)	(120.0,35.093)	(127.5,34.825)	(135.0,34.618)	(142.5,34.465)	(150.0,34.276)	(157.5,34.157)	(165.0,34.032)	(172.5,33.930)	(180.0,33.776)	(187.5,33.684)	(195.0,33.541)	(202.5,33.444)	(210.0,33.316)	(217.5,33.238)	(225.0,33.194)	(232.5,33.076)	(240.0,32.988)
    };
\addplot[color=gray,]
  coordinates {
(7.5,72.982)	(15.0,57.361)	(22.5,52.218)	(30.0,49.368)	(37.5,47.606)	(45.0,46.616)	(52.5,45.673)	(60.0,44.938)	(67.5,44.426)	(75.0,43.976)	(82.5,43.654)	(90.0,43.475)	(97.5,43.481)	(105.0,43.289)	(112.5,42.883)	(120.0,43.142)	(127.5,42.739)	(135.0,42.591)	(142.5,43.252)	(150.0,42.801)	(157.5,43.071)	(165.0,43.768)	(172.5,43.693)	(180.0,43.699)	(187.5,43.501)	(195.0,43.938)	(202.5,43.854)	(210.0,44.268)	(217.5,44.969)	(225.0,44.429)	(232.5,44.265)	(240.0,43.481)
  };
\addplot[color=olive,]
  coordinates {
(7.5,70.083)	(15.0,58.319)	(22.5,54.693)	(30.0,51.934)	(37.5,51.274)	(45.0,49.905)	(52.5,49.077)	(60.0,48.251)	(67.5,48.619)	(75.0,47.443)	(82.5,47.500)	(90.0,46.449)	(97.5,46.739)	(105.0,47.010)	(112.5,46.624)	(120.0,46.208)	(127.5,46.849)	(135.0,45.503)	(142.5,45.754)	(150.0,45.381)	(157.5,45.259)	(165.0,45.361)	(172.5,45.398)	(180.0,45.147)	(187.5,45.185)	(195.0,45.029)	(202.5,44.758)	(210.0,45.223)	(217.5,44.614)	(225.0,45.189)	(232.5,45.062)	(240.0,45.117)
  };
\addplot[color=blue,]
  coordinates {
(7.5,63.603)	(15.0,49.967)	(22.5,45.091)	(30.0,42.408)	(37.5,40.676)	(45.0,39.395)	(52.5,38.430)	(60.0,37.712)	(67.5,37.068)	(75.0,36.525)	(82.5,36.051)	(90.0,35.732)	(97.5,35.318)	(105.0,35.062)	(112.5,34.719)	(120.0,34.437)	(127.5,34.255)	(135.0,34.078)	(142.5,33.924)	(150.0,33.727)	(157.5,33.636)	(165.0,33.461)	(172.5,33.341)	(180.0,33.199)	(187.5,33.119)	(195.0,32.938)	(202.5,32.806)	(210.0,32.753)	(217.5,32.657)	(225.0,32.548)	(232.5,32.435)	(240.0,32.405)
    };
\addplot[color=teal,]
  coordinates {
(6.0,72.378)	(12.0,54.271)	(18.0,48.444)	(24.0,45.183)	(30.0,43.233)	(36.0,41.652)	(42.0,40.575)	(48.0,39.762)	(54.0,39.013)	(60.0,38.415)	(66.0,37.962)	(72.0,37.558)	(78.0,37.184)	(84.0,36.765)	(90.0,36.470)	(96.0,36.132)	(102.0,35.929)	(108.0,35.778)	(114.0,35.516)	(120.0,35.409)	(126.0,35.165)	(132.0,34.936)	(138.0,34.858)	(144.0,34.760)	(150.0,34.691)	(156.0,34.579)	(162.0,34.370)	(168.0,34.363)	(174.0,34.235)	(180.0,34.206)	(186.0,34.134)	(192.0,33.991)	(198.0,33.957)	(204.0,33.783)	(210.0,33.806)	(216.0,33.737)	(222.0,33.608)	(228.0,33.595)	(234.0,33.467)	(238.0,33.364)	(240.0,32.882)
  };
\addplot[color=green,]
  coordinates {
(6.0,73.941)	(12.0,55.584)	(18.0,49.605)	(24.0,46.503)	(30.0,44.625)	(36.0,43.227)	(42.0,42.039)	(48.0,41.233)	(54.0,40.691)	(60.0,40.204)	(66.0,39.474)	(72.0,39.235)	(78.0,38.936)	(84.0,38.689)	(90.0,38.331)	(96.0,38.201)	(102.0,37.863)	(108.0,37.782)	(114.0,37.587)	(120.0,37.271)	(126.0,37.206)	(132.0,37.125)	(138.0,36.984)	(144.0,36.929)	(150.0,36.949)	(156.0,37.045)	(162.0,36.578)	(168.0,36.575)	(174.0,36.655)	(180.0,36.816)	(186.0,36.797)	(192.0,36.699)	(198.0,36.390)	(204.0,36.482)	(210.0,36.512)	(216.0,36.337)	(222.0,36.720)	(228.0,36.412)	(234.0,36.250)	(238.0,36.322)	(240.0,33.711)
  };
  
\addplot[color=red,]
  coordinates {
  (7.5,64.034)	(15.0,50.231)	(22.5,45.568)	(30.0,42.848)	(37.5,41.034)	(45.0,39.817)	(52.5,38.853)	(60.0,38.132)	(67.5,37.490)	(75.0,36.994)	(82.5,36.567)	(90.0,36.166)	(97.5,35.792)	(105.0,35.515)	(112.5,35.287)	(120.0,34.979)	(127.5,34.784)	(135.0,34.600)	(142.5,34.542)	(150.0,34.324)	(157.5,34.105)	(165.0,34.022)	(172.5,33.905)	(180.0,33.769)	(187.5,33.692)	(195.0,33.598)	(202.5,33.423)	(210.0,33.343)	(217.5,33.350)	(225.0,33.227)	(232.5,33.192)	(240.0,33.010)
  };

\addplot[color=pink,]
  coordinates {
  (7.5,67.036)	(15.0,53.382)	(22.5,48.327)	(30.0,45.790)	(37.5,43.926)	(45.0,42.769)	(52.5,41.679)	(60.0,41.094)	(67.5,40.412)	(75.0,40.048)	(82.5,39.523)	(90.0,39.224)	(97.5,38.877)	(105.0,38.498)	(112.5,38.345)	(120.0,38.246)	(127.5,37.947)	(135.0,37.749)	(142.5,37.656)	(150.0,37.624)	(157.5,37.468)	(165.0,37.356)	(172.5,37.171)	(180.0,37.044)	(187.5,37.016)	(195.0,36.878)	(202.5,36.689)	(210.0,36.661)	(217.5,36.520)	(225.0,36.439)	(232.5,36.362)	(240.0,36.361)
  };

\addplot[color=orange,]
  coordinates {
  (7.5,71.253)	(15.0,57.802)	(22.5,55.815)	(30.0,54.383)	(37.5,53.879)	(45.0,51.881)	(52.5,50.137)	(60.0,49.574)	(67.5,49.522)	(75.0,48.463)	(82.5,48.137)	(90.0,47.908)	(97.5,47.268)	(105.0,47.065)	(112.5,46.724)	(120.0,46.714)	(127.5,46.319)	(135.0,46.650)	(142.5,46.426)	(150.0,46.572)	(157.5,46.249)	(165.0,46.713)	(172.5,46.418)	(180.0,46.473)	(187.5,46.248)	(195.0,46.625)	(202.5,45.825)	(210.0,45.806)	(217.5,45.585)	(225.0,45.756)	(232.5,45.597)	(240.0,45.747)
  };
  
\legend{No dropout, Standard $p=0.1$,  Standard $p=0.2$, Standard $p=0.3$, SliceOut $p=0.1$, SliceOut $p=0.2$, SliceOut $p=0.3$, Controlled $p=0.1$,  Controlled $p=0.2$, Controlled $p=0.3$}
\end{axis}
\end{tikzpicture}
\end{center}

\begin{center}
\begin{tikzpicture}
\begin{axis}[
    title={Perplexity vs No. of steps: 2048 embedding dim.},
    xlabel={No. of steps (in 1000s)},
    xmin=0, xmax=300,
    ymin=25, ymax=45,
    xtick={0,50,100,150,200,250, 300},
    ytick={25, 30, 35, 40, 45},
    legend pos=north east,
    ymajorgrids=true,
    grid style=dashed,
    title style={font=\small},
    label style={font=\small},
    legend style={font=\tiny},
]
\addplot[color=black,]
  coordinates {(15.0,39.433)	(30.0,33.967)	(45.0,31.455)	(60.0,30.603)	(75.0,30.643)	(90.0,29.411)	(105.0,28.956)	(120.0,29.503)	(135.0,29.093)	(150.0,28.782)	(165.0,28.380)	(180.0,28.097)	(195.0,28.450)	(210.0,28.806)	(225.0,29.009)	(240.0,28.970)	(255.0,28.596)	(270.0,28.826)	(285.0,28.426) (300.0,28.313)};
\addplot[color=blue,]
  coordinates {(15.0,42.048)	(30.0,36.098)	(45.0,33.427)	(60.0,32.399)	(75.0,32.295)	(90.0,31.114)	(105.0,30.886)	(120.0,30.458)	(135.0,30.019)	(150.0,29.975)	(165.0,29.686)	(180.0,29.715)	(195.0,29.995)	(210.0,29.603)	(225.0,29.294)	(240.0,29.467)	(255.0,29.830)	(270.0,29.416)	(285.0,29.480)
  (300.0,28.12)};
\legend{No dropout, SliceOut $p=0.3$}
\end{axis}
\end{tikzpicture}
\end{center}

\end{document}